\documentclass[11pt]{article}
\usepackage{coling2016}
\usepackage{times}
\usepackage{url}
\usepackage{latexsym}
\usepackage{placeins}
\usepackage{longtable}

\makeatletter
\newcommand{\@BIBLABEL}{\@emptybiblabel}
\newcommand{\@emptybiblabel}[1]{}
\makeatother
\usepackage[colorlinks=true,citecolor=blue,urlcolor=blue,linkcolor=blue,pagebackref=false,hypertexnames=false]{hyperref}

\hypersetup{citecolor=[rgb]{0.078,0.337,0.5},urlcolor=[rgb]{0.078,0.337,0.5},linkcolor=[rgb]{0.078,0.337,0.5}}

\usepackage{graphicx}
\usepackage{caption}
\usepackage{subcaption}
\usepackage{multirow}

\usepackage{amsmath}
\usepackage{amssymb}
\usepackage[parfill]{parskip}

 \usepackage{array}
\newcolumntype{P}[1]{>{\centering\arraybackslash}p{#1}}
  
\newcommand{\refsec}[1]{\hyperref[#1]{section \ref*{#1}}}
\newcommand{\refsecUpper}[1]{\hyperref[#1]{Section \ref*{#1}}}
\newcommand{\reftable}[1]{\hyperref[#1]{Table \ref*{#1}}}
\newcommand{\reffigure}[1]{\hyperref[#1]{Figure \ref*{#1}}}

\newcommand{\numRuns}[0]{50.000}

\title{Optimal Hyperparameters for Deep LSTM-Networks for Sequence Labeling Tasks}

\author{Nils Reimers \and Iryna Gurevych\\
Ubiquitous Knowledge Processing Lab (UKP) and Research Training Group AIPHES\\
Department of Computer Science, Technische Universit\"at Darmstadt\\
Ubiquitous Knowledge Processing Lab (UKP-DIPF) \\ 
German Institute for Educational Research \\
\url{www.ukp.tu-darmstadt.de}
}

\begin{document}
\maketitle
\begin{abstract}
Selecting optimal parameters for a neural network architecture can often make the difference between mediocre and state-of-the-art performance. However, little is published which parameters and design choices should be evaluated or selected making the correct hyperparameter optimization often a ``black art that requires expert experiences'' \cite{Snoek2012}. In this paper, we evaluate the importance of different network design choices and hyperparameters for five common linguistic sequence tagging tasks (POS, Chunking, NER, Entity Recognition, and Event Detection).
We evaluated over \numRuns{} different setups and found, that some parameters, like the pre-trained word embeddings or the last layer of the network, have a large impact on the performance, while other parameters, for example the number of LSTM layers or the number of recurrent units, are of minor importance. We give a recommendation on a configuration that performs well among different tasks. The optimized implementation of our BiLSTM-CRF architecture is publicly available.\footnote{\url{https://github.com/UKPLab/emnlp2017-bilstm-cnn-crf}}

\textit{This publication explains in detail the experimental setup and discusses the results. A condensed version of this paper was presented at EMNLP 2017 \cite{Reimers2017EMNLP}.\footnote{\url{https://arxiv.org/abs/1707.09861}}}
\end{abstract}

\section{Introduction} \label{intro}
Long-Short-Term-Memory (LSTM) Networks \cite{Hochreiter1997} are widely used in NLP and achieve state-of-the-art performance for many sequence-tagging-tasks including Part-of-Speech tagging \cite{Ma2016}, Named Entity Recognition \cite{Ma2016} and Chunking \cite{Sogaard2016}. However, achieving good or even state-of-the-art results with LSTM networks is not straight forward, as it requires the selection and optimization of many hyperparameters, for example tuning the number of recurrent units, the depth of the network, the dropout rate, the pre-trained word embeddings and many more. The selection of hyperparameters  often makes the difference between mediocre and state-of-the-art performance \cite{Hutter2014}.

Besides the already high number of hyperparameters, various extensions to the architecture have been proposed, that add the one or other little knob to the LSTM-network: for example the usage of bidirectional LSTM-layers \cite{Schuster1997}, the usage of variational dropout for recurrent networks \cite{Gal2015}, the usage of a CRF classifier \cite{Huang2015}, the combination of word embeddings with character representations either using CNNs \cite{Ma2016} or LSTMs \cite{Lample2016}, or supervising different tasks at different levels \cite{Sogaard2016}. These extensions can be seen in a broader sense as hyperparameters for the general LSTM architecture for linguistic sequence tagging.

It is commonly agreed that the selection of hyperparameters plays an important role, however, only little research has been published so far to evaluate which hyperparameters and proposed extensions have a high impact on the performance, and which hyperparameters have a rather low impact on the performance. \newcite{Snoek2012} even says that hyperparameter optimization ``... is often a black art that requires expert experience, unwritten rules of thumb, or sometimes brute-force search.'' This situation bears the risk to waste a lot of resources on evaluating irrelevant hyperparameters that have no or minor impact on the performance or to miss out the optimization of important parameters.

The contribution of this paper is an in-depth analysis which hyperparameters are crucial to optimize and which are of less importance. For this, we evaluated over \numRuns{}  BiLSTM networks for five common linguistic sequence tagging tasks. Further, we evaluate different extensions of the BiLSTM network architecture: The BiLSTM-CRF \cite{Huang2015}, the BiLSTM-CNN-CRF \cite{Ma2016} architecture and the BiLSTM-LSTM-CRF architecture \cite{Lample2016}. For each architecture, we ran thousands of different network configurations for the five sequence tagging tasks (POS, Chunking, NER, Entity Recognition, and Event Detection) and measured which configuration gives the best performance. Using the high number of evaluations, we can spot architecture and parameter choices that boost performance in many use cases. 

The main section of this paper is \refsec{sec:evaluation_results}, in which we summarize our results. We show that the pre-trained word embeddings by \newcite{Komninos2016} perform usually the best, that there is no significant difference between the character representation approach by \newcite{Ma2016} and by \newcite{Lample2016}, that the Adam optimizer with Nesterov momentum \cite{Nadam} yields the highest performance and converges the fastest, that while gradient clipping \cite{Mikolov2012} does not help to improve the performance, we observe a large improvement when using gradient normalization \cite{Pascanu2013}, that a BIO tagging scheme is preferred over IOBES and IOB tagging schemes, that adding a CRF-classifier \cite{Huang2015} is helpful, that the variational dropout of \newcite{Gal2015} should be applied to output and the recurrent dimensions, that two stacked recurrent layers usually performs best, and that the impact of the number of recurrent units is rather small and that around 100 recurrent units per LSTM-network appear to be a good rule of thumb.

\refsecUpper{sec:mtl} summarizes the results of our Multi-Task Learning experiments. We show that multi-task learning is beneficial only in certain circumstances, when the jointly learned tasks are linguistically similar. In all other cases, using a single task learning setup is the better option. Our results show, that multi-task learning is especially sensitive to the selected hyperparameters. In contrast to previous approaches in this field, our results show that it is beneficial to have besides shared LSTM-layers also task-dependent LSTM-layers that are optimized for each task.

\section{Related Work}
A well-known method for hyperparameter search is \textit{grid search}, also called \textit{parameter sweeping}. It is an exhaustive search through a manually defined subset of possible hyperparameters. However, the set of to-test parameters can grow quite quickly, especially as different combinations of hyperparameters must be tested. For our experiments we tuned 11 different hyperparameters. Even though we limit the choices per hyperparameter to fairly small, manually selected sets, we would need to evaluate more than 6 Million combinations per task if we wanted to test all combinations.

Since grid search is a computationally expensive approach, several alternatives have been proposed. \textit{Randomized search} samples parameter settings at random for a fixed number of times. \newcite{Bergstra2012} show empirically and theoretically that randomly chosen trials are more efficient for hyperparameter optimization than trials on a grid. They find that neural networks configured by randomized search is able to find models that are ``as good or better within a small fraction of the computation time'' compared to networks configured by pure grid search. Granting randomized search the same computational budget it finds better models by effectively searching a larger configuration space. \newcite{Bergstra2012} show that for most datasets only a few of the hyperparameters really matter. This makes grid search a poor choice, as a lower number of parameter settings for the important factors are tested. 
In contrast to grid search, randomized search is also easier to carry out: The search can be stopped, resumed or modified at any time without jeopardizing the experiment.

Randomized search has the disadvantage that it does not adapt its behavior based on previous outcomes. In some cases, a single poorly chosen hyperparameter, for example a far too large learning rate or a far too large dropout rate, will prevent the model from learning effectively.  For example if the learning rate must be in the range between 0.01 and 0.1 to yield good results, but values between 0 and 1 are tested by randomized search, then over 90\% of all trials will fail due to a badly chosen learning rate. This requires carefully selecting the options and ranges for the optimized hyperparameters when applying randomized search.

To avoid this, Bayesian Optimization methods \cite{Snoek2012} are able to learn from the training history and give a better estimation for the next set of parameters. A Bayesian optimization method consists of developing a statistical model between the hyperparameters and the objective functions. It makes the assumption that there is a smooth but noisy function that maps between the hyperparameters and the objective function. \newcite{Bergstra2013} found that a Tree of Parzen Estimators (TPE) \cite{Bergstra2011} was able to find the best configuration for the evaluated datasets and it required only a small fraction of the time that they allocated to randomized search. However, in comparison to randomized search, Bayesian optimization methods require the understanding and implementation of additional complexity, which makes it a less popular choice for many researchers. 

\textit{Ad hoc manual tuning} is still a commonly and often surprisingly effective approach for hyperparameter tuning \cite{Hutter2015}. The algorithm inventor iteratively selects different architectures and hyperparameters and homes in to a high-performance region of the hyperparameter space. Coarse grained grid-search and randomized search can be helpful to identify those high-performance regions.

Even though that it is widely recognized that the network architecture and the selected hyperparameters are crucial and adding little knobs like variational dropout \cite{Gal2015}, a CRF-classifier \cite{Huang2015} or adding character-based representations \cite{Ma2016,Lample2016} can significantly change the performance of the model, only little is reported which knobs are the most important to tune. \newcite{Hutter2014} describe a method to gain insights into the relative importance of hyperparameters by using random forest predictions with functional ANOVA compositions \cite{Gramacy2013}. However, it does not help to prioritize which parameter choices and extensions to implement in the first place. More practical recommendations for training deep neural network architectures and selecting hyperparameters are given by \newcite{Bengio2012}.

\FloatBarrier
\section{LSTM-Networks for Sequence Tagging}\label{sec:lstm_model}
\begin{figure}[ht]
\centering
  \includegraphics[width=.85\textwidth]{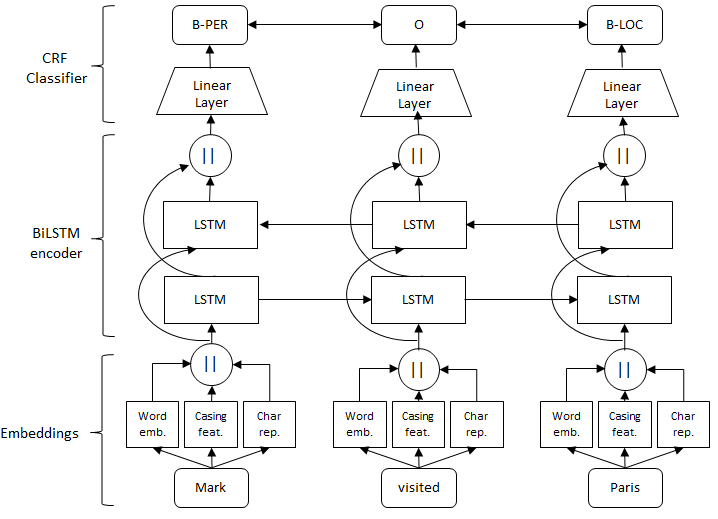}
\caption{Architecture of the BiLSTM network with a CRF-classifier. A fixed sized character-based representation is derived either with a Convolutional Neural Network or with a BiLSTM network. }
\label{fig:bilstm_network}
\end{figure}

LSTM-Networks are a popular choice for linguistic sequence tagging and show a strong performance in many tasks. \reffigure{fig:bilstm_network} shows the principle architecture of a BiLSTM-model for sequence tagging. A detailed explanation of the model can be found in \cite{Huang2015,Ma2016,Lample2016}.

Each word in a sentence is mapped to a (pre-trained) word embedding. As word embeddings are usually only provided for lower cased word, we add a capitalization feature that captures the original casing of the word. The capitalization feature assigns the label \textit{numeric}, if each character is numeric, \textit{mainly numeric} if more than 50\% of the characters are numeric, \textit{all lower} and \textit{all upper} if all characters are lower/upper cased, \textit{initial upper} if the initial character is upper cased, \textit{contains digit} if it contains a digit and \textit{other} if none of the previous rules applies. The capitalization feature is mapped to a seven dimensional one-hot vector. 

\begin{figure}[ht]
\centering
\begin{subfigure}{.45\textwidth}
  \centering
  \includegraphics[width=\linewidth]{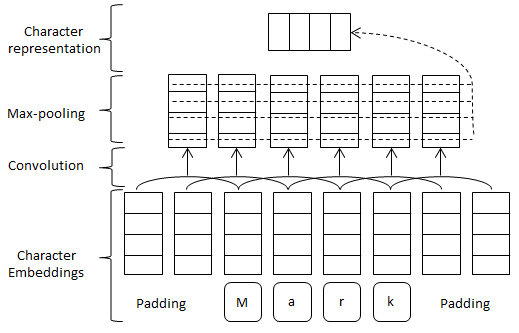}
  \subcaption{CNN approach}
\end{subfigure}
\begin{subfigure}{.45\textwidth}
  \centering
  \includegraphics[width=\linewidth]{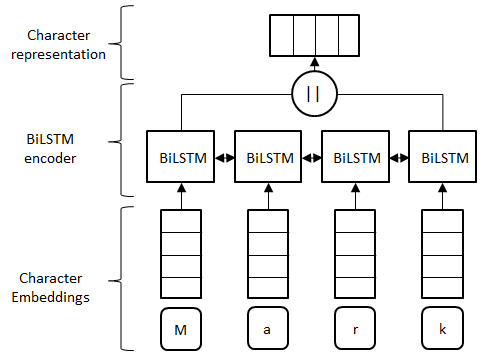}
  \subcaption{BiLSTM approach}
\end{subfigure}
\caption{(a) Character-based representation using a convolutional neural network \protect\cite{Ma2016}, (b) character-based representation using BiLSTM-networds \protect\cite{Lample2016}. }
\label{fig:character_representation}
\end{figure}

We added the option to  derive a fixed-size dense representation based on the characters of word. The process is depicted in \reffigure{fig:character_representation}. Each character in a word is mapped to a randomly initialized 30-dimensional embedding. For the CNN approach by \newcite{Ma2016}, a convolution with 30 filters and filter length of 3 (i.e.\ character trigrams) is used, followed by a max-over-time pooling. For the BiLSTM approach by \newcite{Lample2016}, the character embeddings are fed into a BiLSTM encoder, each LSTM-network with 25 recurrent units. The last output of the two LSTM-networks are then combined to form a 50 dimensional character-based representation.

The word embedding, the capitalization feature and the character-based representation are concatenated ($\parallel$) and are used for a BiLSTM-encoder. One LSTM network runs from the beginning of the sentence to the end while the other runs in reverse. The output of both LSTM networks are concatenated and are used as input for a classifier. For a Softmax classifier, we map the output through a dense layer with softmax as activation function. This gives for each token in the sentence a probability distribution for the possible tags. The tag with the highest probability is selected. For the CRF-classifier, the concatenated output is mapped with a dense layer and a linear activation function to the number of tags. Then, a linear-chain Conditional Random Field maximizes the tag probability of the complete sentence.

\FloatBarrier
\section{Benchmark Tasks}\label{sec:benchmark_tasks}
In this section, we briefly introduce the used tasks to evaluate the different hyperparameter choices for deep LSTM networks. We use five classical NLP tasks: Part-of-Speech tagging (POS), chunking (Chunking), Named Entity Recognition (NER), entity recognition (Entities), and event detection (Events). \reftable{table:datasets_overview} gives an overview of the used datasets.

\begin{table}[h]
\centering
\begin{tabular}{|c|c|c|c|c|}
\hline
\textbf{Task} & \textbf{Dataset} & \textbf{Training sentences} & \textbf{Test sentences} & \textbf{\#tags} \\ \hline
POS & WSJ & 500 & 5459 & 45 \\ \hline
Chunking & ConLL 2000 (WSJ) & 8926 & 2009 & 23 \\ \hline
NER &  CoNLL 2003 (Reuters) & 13862 & 3420 & 9 \\ \hline
Entities &  ACE 2005 & 15185 & 674 & 15\\ \hline
Events & TempEval3 &  4090 & 279 & 3\\ \hline
\end{tabular}
\caption{Overview of the benchmark tasks. For the POS task, we only use the first 500 sentences of the training set. The number of tags for Chunking, NER, Entities and Events includes a BIO tagging scheme.  }
\label{table:datasets_overview}
\end{table}

\textbf{Part-of-Speech tagging.} Part-of-Speech tagging aims at labeling each token in the text with a tag indicating its syntactic role, e.g.\ noun, verb etc. The typical benchmark setup is described in detail in \cite{Toutanova_2003}. Usually, the sections 0-18 of the Wall Street Journal (WSJ) are used for training, while sections 19-21 are used for validation and hyperparameter optimization and sections 22-24 are used for testing.

Part-of-Speech tagging is a relatively simple task. \newcite{Toutanova_2003} report an accuracy of 97.24\% on the test set. The estimated error rate for the PennTree Bank POS information is approximately 3\% \cite{Marcus_1993}. \newcite{Marcus_1993} found in an early experiment on the Brown corpus that the disagreement between two annotators when correcting the output of an automatic tagger is 4.1\% and 3.5\% once one difficult text is excluded. An improvement lower than the error rate, i.e.\ above 97\% accuracy, is unlikely to be meaningful and has a high risk of being the result of chance.

In order to be able to compare different neural architectures and hyperparameters for Part-of-Speech tagging, we increased the difficulty of the task by decreasing the training set. We decided to decrease the size of the training set to the first 500 sentences in the Wall Street Journal. The development and test sets were kept unchanged. This decreased the accuracy to a range of about 94-95\%, fairly below the (estimated) upper-bound of 97\%.

\textbf{Chunking.} Chunking aims at labeling segments of a sentence with syntactic constituents such as noun or verb phrase. Each word is assigned a single tag. To note the beginning of a new segment, a tagging scheme, for example a BIO tagging scheme, can be used. Here, the tag \texttt{B-NP} denotes the beginning of a noun phrase and \texttt{I-NP} would denote each other word of the noun phrase. We evaluate our systems using the CoNLL 2000 shared task\footnote{\url{http://www.cnts.ua.ac.be/conll2000/chunking/}}. Sections 15-18 of the Wall Street Journal are used for training, section 19 is used as development set, and section 20 is used for testing. The performance is computed using the $F_1$ score, which is the harmonic mean of the precision and recall. Note that besides the class label the span of the segment must perfectly match the gold data. If a produced segment is a single token too long or to short, it is considered an error. Current state-of-the-art performance for this setup is at about 95\% $F_1$-measure \cite{Hashimoto_2016}.

\textbf{NER.} Named Entity Recognition (NER) aims at labeling named entities, like person names, locations, or company names, in a sentence. As in the chunking task, a named entity can consist of several tokens and a tagging scheme is involved to denote the beginning and the end of an entity. In this paper we use the CoNLL 2003 setup\footnote{\url{http://www.cnts.ua.ac.be/conll2003/ner/}} which provides train, development and test data, as well as an evaluation script. Named entities are categorized in the four categories \textit{person names}, \textit{locations}, \textit{organizations} and \textit{miscellaneous}. The performance is computed using the $F_1$ score for all named entities. Current state-of-the-art performance for this setup is at 91.21\% $F_1$-measure \cite{Ma2016}.

\textbf{Entities.} In contrast to the CoNLL 2003 NER dataset, the ACE 2005 dataset\footnote{\url{https://catalog.ldc.upenn.edu/LDC2006T06}} annotated not only named entities, but all words referring to an entity, for example the words \textit{U.S. president}. The entities are categorized in the seven categories: \textit{persons}, \textit{organizations}, \textit{geographical/social/political entities}, \textit{locations}, \textit{facilities}, \textit{vehicle} and \textit{weapon}. We use the same data split as \newcite{Li_2013} partitioning the data into 529 train documents, 40 development documents, and 30 test documents. For evaluation, we compute the $F_1$ score as before.

\textbf{Events.} The TempEval3 Task B\footnote{\url{https://www.cs.york.ac.uk/semeval-2013/task1/}} \cite{tempeval3} defines an \textit{event} as a cover term for situations that \textit{happen} or \textit{occur} \cite{timeml_annotation_guidelines}. The smallest extent of text, usually a single word, that expresses the occurrence of an event, is annotated. In the following example, the words \textit{sent} and \textit{spent} expresses events.
\begin{quote}
\textit{He was \textbf{sent} into space on May 26, 1980. He \textbf{spent} six days aboard the Salyut 6 spacecraft.}
\end{quote}
The performance is measured as for the chunking, NER and Entities task by computing the $F_1$ score. The best system of the TempEval3 shared task achieved a $F_1$ score of 81.05\% \cite{tempeval3}.

\FloatBarrier
\section{Evaluation Methology}
In this paper we evaluate different hyperparameters and variants of the LSTM sequence tagging architecture on five common NLP tasks: Part-of-Speech tagging, chunking, NER, entity recognition and event recognition. However, the goal of this paper is \textit{not} to find one specific configuration that performs best on these tasks. As shown in our publication \cite{Reimers2017EMNLP}, presenting a single configuration together with the achieved performance is not meaningful. The seed value of the random number generator has statistically significant impact on the outcome of a single training run. The difference in terms of performance can be as large as 8.23 percentage points in $F_1$-score for the same network with the same hyperparameters when it is trained twice with different seed values. An overview is shown in \reftable{table:random_init}. 

\begin{table}[h]
\centering
\begin{tabular}{|c|c|c|c|c|c|}
\hline
\textbf{Task} & \textbf{Dataset} & \textbf{\# Configs} & \textbf{Median Difference} & \textbf{95th percentile} & \textbf{Max. Difference}   \\ \hline
POS & Penn Treebank & 269 & 0.17\% & 0.78\% & 1.55\% \\ \hline
Chunking & CoNLL 2000 & 385 & 0.17\% & 0.50\% & 0.81\% \\ \hline
NER & CoNLL 2003 & 406 & 0.38\% & 1.08\% & 2.59\% \\ \hline
Entities & ACE 2005 & 405 & 0.72\% & 2.10\% & 8.23\% \\ \hline
Events & TempEval 3  & 365 & 0.43\% & 1.23\% & 1.73\% \\ \hline
\end{tabular}
\caption{The table depicts the median, the 95th percentile and the maximum difference between networks with the same hyperparameters but different random seed values.}
\label{table:random_init}
\end{table}

In this publication, we are interested to find design choices that perform \textbf{robustly}, i.e.\ that perform well independent of the selected hyperparameters and independent of the sequence of random numbers. Such design choices are especially interesting when the BiLSTM architecture is  applied to  new tasks, new domains or new languages. In order to find the most robust design option, e.g.\ to decide whether a Softmax classifier or a CRF classifier as last layer is better, we sampled randomly a large number of network configurations and evaluated each configuration with a Softmax classifier as well as with a CRF classifier as last layer. 

\begin{figure}[ht]
\centering
  \includegraphics[width=.4\textwidth]{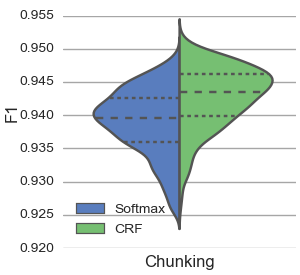}
\caption{Probability density function for the chunking task using a Softmax classifier or a CRF classifier as a last layer of a BiLSTM-network. The median and the quartiles are plotted as dashed lines. }
\label{fig:example_violin_plot}
\end{figure}

The achieved test performances can be plotted in a violin plot as shown in \reffigure{fig:example_violin_plot}. The violin plot is similar to a boxplot, however, it estimates from the samples the probability density function and depicts it along the Y-axis. If a violin plot is wide at a certain location, then achieving this test performance is especially likely. Besides the probability density it also shows the median as well as the quartiles. In \reffigure{fig:example_violin_plot} we can observe that a CRF-classifier usually results in a higher performance than a softmax classifier for the chunking dataset. As we sample and run several hundred hyperparameter configurations, random noise, e.g.\ from the weight initialization, will cancel out and we can conclude that the CRF-classifier is a better option for this task.

For brevity reasons, we show the violin plot only in certain situations. In most cases, we report a table like \reftable{table:demo}. The table shows how many network configurations were sampled randomly for each task with the parameters described in \refsec{sec:evaluated_hyperparameters}. For the Chunking task 229 configurations were sampled randomly. Each configuration was evaluated with a Softmax classifier as well as a CRF classifier as a last layer of the network. For 219 of the 230 configurations (95.2\%), the CRF setup achieved a better test performance than the setup with a Softmax classifier. The average depicted at the bottom of the table is the macro average about how often each evaluated option achieved the best performance for the five tasks.

Besides measuring which option achieves the better performance, we also compute the median of the differences to the best option. For the Chunking task, the option that resulted in most cases in the best performance was the CRF classifier. Hence, for the Softmax option we compute the median difference of test performance to the best option. Let $S_i$ be the test performance ($F_1$-measure) for the Softmax setup for configuration $i$ and $C_i$ the test performance for the CRF setup. We then compute $\Delta F_1 = \text{median}(S_1 - C_1, S_2 - C_2, \dots , S_{230}-C_{230})$. For the chunking task, the median difference was $\Delta F_1 = -0.38\%$, i.e.\ the setup with a Softmax classifier achieved on average an $F_1$-score of 0.38 percentage points below that of the CRF setup.

\begin{table}[h]
\centering
\begin{tabular}{|c|c|c|c|c|}
\hline
\textbf{Task} & \textbf{\# Configs} & \textbf{Softmax} & \textbf{CRF}   \\ \hline
POS & 114 & 19.3\% & \textbf{80.7\%}$\dagger$ \\
$\Delta Acc.$ &  & -0.19\% &  \\  
$\sigma$ &  & 0.0149 & 0.0132 \\ \hline 
Chunking & 230 & 4.8\% & \textbf{95.2\%}$\dagger$ \\
$\Delta F_1$ &  & -0.38\% &  \\  
$\sigma$ &  & 0.0058 & 0.0051 \\ \hline 
NER & 235 & 9.4\% & \textbf{90.6\%}$\dagger$ \\
$\Delta F_1$ &  & -0.67\% &  \\  
$\sigma$ &  & 0.0081 & 0.0060$\dagger$ \\ \hline 
Entities & 214 & 13.1\% & \textbf{86.9\%}$\dagger$ \\
$\Delta F_1$ &  & -0.85\% &  \\  
$\sigma$ &  & 0.0157 & 0.0140 \\ \hline 
Events & 203 & \textbf{61.6\%}$\dagger$ & 38.4\% \\
$\Delta F_1$ &  &  & -0.15\% \\  
$\sigma$ &  & 0.0052 & 0.0057 \\ \hline 
\hline
Average &  & 21.6\% & \textbf{78.4\%} \\ \hline 
\end{tabular}
\caption{Network configurations were sampled randomly and each was evaluated with each classifier as a last layer. The first number in a cell depicts in how many cases each classifier produced better results than the others. The second number shows the median difference to the best option for each task. Statistically significant differences with $p < 0.01$ are marked with $\dagger$.}
\label{table:demo}
\end{table}

Based on the outcome of the different runs, we use a two-sided binomial test to find options that perform \textit{statistically significant} better than other options. As threshold we use $p < 0.01$. We compute the standard deviation $\sigma$ of the achieved performance scores to measure the dependence on the remaining configuration of the network and / or the random seed value. We use a Brown-Forsythe test with threshold $p < 0.01$ to identify standard deviations that are statistically significant from others. The best result and all statistically equal results are marked with a $\dagger$. If none of the options in a row has a $\dagger$, then we did not observe a statistically significant difference between the options.

\FloatBarrier

\section{Experimental Setup}\label{sec:experimental_setup}
We implemented the BiLSTM networks using Keras\footnote{\url{http://keras.io}} version 1.2.2. The source code for all our experiments as well a spreadsheet containing the results of the \numRuns{} different runs for further statistical analysis can be found on GitHub\footnote{\url{https://github.com/UKPLab/emnlp2017-bilstm-cnn-crf}}. 

If a token does not appear in the vocabulary of the pre-trained word embeddings, we performed some normalization steps: We converted the token to lower-case and replaced numbers by a special \texttt{NUMBER} token. If it is still not in the vocabulary, and it appears at least 50 times in the training set, the token is added to the vocabulary and a random word embedding is assigned. If it appears less than 50 times, it is replaced by a special  \texttt{UNKNOWN} token. 

We used a mini-batch size of 32 for training and applied an early stopping if the development score does not increase for more than 5 training epochs. The test score is then taken from the run with the highest development score.

\subsection{Evaluated Hyperparameters}\label{sec:evaluated_hyperparameters}
We evaluate the following hyperparameters.

\textbf{Pre-trained Word Embeddings.} Representing words as dense vectors, usually with 100 - 300 dimensions, is a widely used technique in NLP and it can significantly increase the performance \newcite{Collobert_2011}. Word embeddings provide a good generalization to unseen words since they can capture general syntactic as well as semantic properties of words. Which of the many published word embedding generation processes results in the best embeddings however is unclear and depends on many factors, including the dataset from which they are created and for which purpose they will be used. For our five benchmark datasets we evaluate different popular, publicly available pre-trained word embeddings. We evaluate the \textbf{GoogleNews} embeddings\footnote{\url{https://code.google.com/archive/p/word2vec/}} trained on part of the Google News dataset (about 100 billion words) from \newcite{word2vec}, the Bag of Words (\textbf{Levy BoW}) as well as the dependency based embeddings (\textbf{Levy Dep.})\footnote{\url{https://levyomer.wordpress.com/2014/04/25/dependency-based-word-embeddings/}} by \newcite{Levy_dep_embeddings} trained on the English Wikipedia, three different \textbf{GloVe} embeddings\footnote{\url{http://nlp.stanford.edu/projects/glove/}} from \newcite{glove} trained either on Wikipedia 2014 + Gigaword 5 (about 6 billion tokens) or on Common Crawl (about 840 billion tokens), and the \textbf{\newcite{Komninos2016}} embeddings\footnote{\url{https://www.cs.york.ac.uk/nlp/extvec/}} trained on the Wikipedia August 2015 dump (about 2 billion tokens). We also evaluate the approach of \textbf{Fasttext} (FastText), which does not train word embeddings directly, but trains embeddings for n-grams with length 3 to 6. The embedding for a word is then defined as the sum of the embeddings of the n-grams. This allows deriving a meaningful embedding even for rare words, which are often not part of the vocabulary for the other pre-trained embeddings. As the results in \refsec{sec:word_embeddings} show, the different word embeddings lead to significant performance differences.

\textbf{Character Representation.} Character-level information, especially pre- and suffixes of words, can contain valuable information for linguistic sequence labeling tasks like Part-of-Speech tagging. However, instead of hand-engineered features, \newcite{Ma2016} and \newcite{Lample2016} present a method to learn task-specific character level representations while training. \newcite{Ma2016} use convolutional neural networks (\textbf{CNNs}) \cite{LeCun1989} to encode the trigrams of a word to a fixed-sized character-based representation. On the other hand, \newcite{Lample2016} use bidirectional \textbf{LSTMs} to derive the character-based representation. In our experiments, we evaluate both approaches with the parameters mentioned in the respective papers.

\textbf{Optimizer.} The optimizer is responsible of the minimization of the objective function of the neural network. A commonly selected optimizer is stochastic gradient descent (\textbf{SGD}), which proved itself as an efficient and effective optimization method for a large number of published machine learning systems. However, SGD can be quite sensitive towards the selection of the learning rate. Choosing a too large rate can cause the system to diverge in terms of the objective function, and choosing a too low rate results in a slow learning process. Further, SGD has troubles to navigate ravines and at saddle points. To eliminate the short comings of SGD, other gradient-based optimization algorithms have been proposed. Namely \textbf{Adagrad} \cite{Adagrad}, \textbf{Adadelta} \cite{Adadelta}, \textbf{RMSProp} \cite{RMSProp}, \textbf{Adam} \cite{Adam}, and \textbf{Nadam} \cite{Nadam}, an Adam variant that incorporates Nesterov momentum \cite{Nesterov:1983}. The results can be found in \refsec{sec:optimizers}.

\textbf{Gradient Clipping and Normalization.} Two widely known issues with properly training recurrent neural networks are the \textit{vanishing} and the \textit{exploding} gradient problem \cite{Bengio1994}. While the vanishing gradient problem is countered by using LSTM networks, exploding gradients, i.e.\ gradients with extremely large values, are still an issue. Two common strategies to deal with the exploding gradient problem are \textbf{gradient clipping} \cite{Mikolov2012} and \textbf{gradient normalization} \cite{Pascanu2013}. Gradient clipping involves clipping the gradient's components element-wise if it exceeds a defined threshold $\tau$. For each gradient component we compute $\hat g_{ij} = \max(-\tau,\min(\tau, g_{ij}))$. The matrix $\hat g$ is then used for the weight update. Gradient normalization has a better theoretical justification and rescales the gradient whenever the norm $||g||_2$ goes over a threshold $\tau$: $\hat g = (\tau/||g||_2) g \text{ if } ||g||_2 > \tau$. In \refsec{sec:gradient_clipping} we evaluate both approaches and their importance for a good test performance.

\textbf{Tagging schemes.} While the POS task assigns a syntactic role for each word in a sentence, the remaining four tasks associate labels for segments in a sentence. This is achieved by using a special tagging scheme to identify the segment boundaries. We evaluate the \textbf{BIO}, \textbf{IOB}, and \textbf{IOBES} schemes. The \textit{BIO} scheme marks the beginning of a segment with a \texttt{B-} tag and all other tokens of the same span with a \texttt{I-} tag. The \texttt{O} tag is used to tokens that are outside of a segment. The \textit{IOB} scheme is similar to the BIO scheme, however, here the tag \texttt{B-} is only used to start a segment if the previous token is of the same class but is not part of the segment. The chunking data from CoNLL 2000 is provided using the BIO-scheme, while the NER dataset from CoNLL 2003 has an IOB tagging scheme. The \textit{IOBES} scheme distinguishes between single token segments, which are tagged with an \texttt{S-} tag, the beginning of a segment that is tagged with a \texttt{B-} tag, the last token of a segment which is tagged with an \texttt{E-} tag, and tokens inside a segment which are tagged with an \texttt{I-} tag. It is unclear which tagging scheme is in general better. \newcite{Collobert_2011} decided to use the most expressive IOBES tagging scheme for all their tasks. The results of our evaluations can be found in \refsec{sec:tagging_schemes}.

Note, when a tagging scheme is used, the classifier might produce invalid tags, for example an \texttt{I-} tag without a previous \texttt{B-} tag to start the segment. We observed a high number of invalid tags especially for the softmax classifier which does not take the label sequence into account, while for the CRF-classifier invalid tags occurred only rarely. Depending on the used evaluation script such invalid tags might result into an erroneous computation of the $F_1$-score. To eliminate invalid tags, we applied two post-processing strategies: Either set all invalid tags to \texttt{O} or change the tag to \texttt{B-} to start a new segment. Usually setting all invalid tags to \texttt{O} resulted in slightly superior results. 

\textbf{Classifier.} We evaluated two options for the last layer of the network. The first option is a dense layer with softmax activation function, i.e.\ a \textbf{softmax classifier}. This classifier produces a probability distribution for the different tags for each token in the sentence. In this approach, each token in a sentence is considered independently and correlations between tags in a sentence cannot be taken into account. As second option, we evaluate a dense layer with a linear activation function followed by a linear-chain Conditional Random Field (CRF). We call this variant \textbf{CRF classifier}. This option is able to maximize the tag probability of the complete sentence. This is especially helpful for tasks with strong dependencies between token tags, for example if one of the above described tagging schemes is used where certain tags cannot follow other tags. This approach is also known as BiLSTM-CRF \cite{Huang2015}. The results can be found in \refsec{sec:classifiers}.

\textbf{Dropout.} Dropout is a popular method to deal with overfitting for neural networks \cite{Srivastava2014}. In our setup, we evaluate three options: \textbf{No dropout}, \textbf{naive dropout}, and \textbf{variational dropout}. \textit{Naive dropout} is the simplest form of dropout: We apply a randomly selected dropout mask for each LSTM-output. The mask changes from time step to time step. The recurrent connections are not dropped. As noted by \newcite{Gal2015}, this form of dropout is suboptimal for recurrent neural networks. They propose to use the same dropout mask for all time steps of the recurrent layer, i.e.\ at each time step the same positions of the output are dropped. They also propose to use the same strategy to drop the recurrent connections. This approach is known as \textit{variational dropout}. Further details can be found in \cite{Gal2015}. The fraction $p$ of dropped dimensions is a hyperparameter and is selected randomly from the set $\{\boldsymbol{0.0, 0.05, 0.1, 0.25, 0.5}\}$. Note, for variational dropout, the fraction $p$ is selected independently for the output units as well as for the recurrent units. The results can be found in \refsec{sec:dropout}.

\textbf{Number of LSTM-Layers.} We evaluated \textbf{1}, \textbf{2}, and \textbf{3} stacked BiLSTM-layers. The results can be found in \refsec{sec:num_lstm_layers}.

\textbf{Number of Recurrent Units.} The number of recurrent units was selected from the set $\{\boldsymbol{25, 50, 75, 100, 125}\}$. The forward and reverse running LSTM-networks had the same number of recurrent units. In case of multiple layers, we selected for each BiLSTM-layer a new value. Increasing layers sizes were forbidden. For \refsec{sec:num_lstm_layers} we also evaluated networks with in $60 \leq u \leq 300$ recurrent units.

\textbf{Mini-batch Size.} We evaluated mini-batch sizes of 1, 8, 16, 32, and 64 sentences.

\textbf{Backend.} Keras offers the option to choose either \textbf{Theano} or \textbf{Tensorflow} as backend. Due to slightly different implementations of the mathematical operations and numerical instabilities, the results can differ between Theano and Tensorflow. However, as shown in \refsec{sec:theano_vs_tensorflow}, both result in approximately the same test performances. The versions 0.8.2 for Theano and 0.12.1 for Tensorflow were used.

\section{Evaluation Results}\label{sec:evaluation_results}

The following table gives an overview of the results of our experiments. Details can be found in the consecutive subsections.

\begin{longtable}{|P{3cm}|P{3cm}|P{1.5cm}|p{7cm}|}
\hline
\textbf{Parameter} &  \textbf{Recom. Config} & \textbf{Impact} & \textbf{Comment}     \\ \hline
Word Embeddings & Komninos et al. & High & The embeddings by \newcite{Komninos2016} resulted for the most tasks in the best performance. For the POS tagging task, the median difference to e.g.\ the GloVe embeddings was 4.97 percentage points. The GloVe embeddings trained on Common Crawl were especially well suited for the NER task, due to their high coverage. More details in \refsec{sec:word_embeddings}. \\ \hline

Character Representation & CNNs \cite{Ma2016}  & Low - Medium & Character-based representations were in a lot of tested configurations not that helpful and could not improve the performance of the network. The CNN approach by \newcite{Ma2016} and the LSTM approach by \newcite{Lample2016} performed on-par. The CNN approach should be preferred due to the higher computational efficiency. More details in \refsec{sec:char_representation}. \\ \hline

Optimizer & Nadam & High & Adam and Adam with Nesterov momentum (Nadam) usually performed the best, followed by RMSProp. Adadelta and Adagrad had a much higher variance in terms of test performance and resulted on average to far worse results. SGD failed in a high number of cases to converge to a minimum, likely due to its high sensitivity of the learning rate. Nadam was the fastest optimizer. More details in \refsec{sec:optimizers}. \\ \hline

Gradient Clipping / Normalization & Gradient Normalization with $\tau=1$ & High & Gradient clipping does not improve the performance. Gradient normalization as described by \newcite{Pascanu2013} improves significantly the performance with an observed average improvement between 0.45 and 0.82 percentage points. The threshold $\tau$ is of minor importance, with $\tau=1$ giving usually the best results. More details in \refsec{sec:gradient_clipping}.   \\ \hline

Tagging Scheme & BIO & Medium & The BIO and IOBES tagging scheme performed consistently better than the IOB tagging scheme. IOBES does not give a significant performance increase compared to the BIO tagging scheme. More details in \refsec{sec:tagging_schemes}. \\ \hline

Classifier & CRF & High & Using a CRF instead of a softmax classifier as a last layer gave a large performance increase for tasks with a high dependency between tags. This was also true for stacked BiLSTM layers. For tasks without dependencies between the tags, Softmax performed better. More details in \refsec{sec:classifiers}. \\ \hline

Dropout & Variational & High & Variational dropout \cite{Gal2015} outperformed significantly no dropout and also naive dropout. The best result was achieved, when dropout was applied both to the output units as well as to the recurrent units of the LSTM networks. More details in \refsec{sec:dropout}. \\ \hline

\#LSTM Layers & 2 & Medium & If the number of recurrent units is kept constant, two stacked BiLSTM-layers resulted in the best performance. More details in \refsec{sec:num_lstm_layers}. \\ \hline

Recurrent Units & 100 & Low & The number of recurrent units, as long as it is not far too large or far too small, has only a minor effect on the results. A value of about 100 for each LSTM-network appears to be a good rule of thumb for the tested tasks. More details in \refsec{sec:num_recurrent_units}.  \\ \hline

Mini-batch Size & 1-32 & Medium & The optimal size for the mini-batch appears to depend on the task. For POS tagging and event recognition, a size of 1 was optimal, for chunking a size of 8 and for NER and Entity Recognition a size of 31. More details in \refsec{sec:mini-batch-size}. \\ \hline

Backend & - & None & Theano as well as Tensorflow performed equally in terms of test performance. The selection of the backend should therefore depend on other criteria, e.g.\ on run time. More details in \refsec{sec:theano_vs_tensorflow}. \\ \hline
\end{longtable}

\FloatBarrier
\subsection{Word Embeddings} \label{sec:word_embeddings}
\reftable{table:word_embeddings} shows the impact of different pre-trained word embeddings on the five evaluated benchmark tasks. The embeddings by \newcite{Komninos2016} give the best performance on all tasks except for the CoNLL 2003 NER and CoNLL 2000 chunking tasks, where they perform on-par with the GloVe embeddings on CommonCrawl and the \newcite{Levy_dep_embeddings} dependency-based embeddings. The median difference in test performance is quite large. For example for the POS tagging task, the embeddings by \newcite{Komninos2016} give on average a 4.97 percentage points higher accuracy than the GloVe2 embeddings (100 dimensions, trained on Wikipedia and Gigaword).

The only datasets where the \newcite{Komninos2016} embeddings would not necessarily be the best selection is for Chunking and the NER task. For Chunking, the dependency based embeddings by \newcite{Levy_dep_embeddings} gave in most cases the best performance. However, this difference is statistically not significant (p=6.5\%) and looking at the mean performance shows that both embeddings are on-par.

For the NER task, the GloVe embeddings trained on 840 billion tokens from Common Crawl (GloVe 3) resulted in most cases in the best performance. As before, the difference to the \newcite{Komninos2016} embeddings is statistically insignificant (p=33.0\%) and more runs would be required to determine which embeddings would be the best selection.

The n-gram embedding approach FastText by \newcite{Fasttext}, which allows deriving meaningful word embeddings also for rare words which are often not in the vocabulary for the other approaches, does not yield a good performance in any of the benchmark tasks.

\begin{table}[h]
\centering
\begin{tabular}{|c|c|c|c|c|c|c|c|c|}
\hline
\textbf{Dataset} &  \textbf{Le. Dep.} & \textbf{Le. BoW} & \textbf{GloVe1} & \textbf{GloVe2} & \textbf{GloVe3} & \textbf{Komn.} & \textbf{G. News} & \textbf{FastText}  \\ \hline
POS  & 6.5\% & 0.0\% & 0.0\% & 0.0\% & 0.0\% & \textbf{93.5\%}$\dagger$ & 0.0\% & 0.0\% \\
$\Delta Acc.$   & -0.39\% & -2.52\% & -4.14\% & -4.97\% & -2.60\% &  & -1.95\% & -2.28\% \\  
$\sigma$  & 0.0125$\dagger$ & 0.0147 & 0.0203 & 0.0136 & 0.0097 & 0.0058$\dagger$ & 0.0118 & 0.0120 \\ \hline 
Chunking  & \textbf{60.8\%}$\dagger$ & 0.0\% & 0.0\% & 0.0\% & 0.0\% & 37.1\%$\dagger$ & 2.1\% & 0.0\% \\
$\Delta F_1$   &  & -0.52\% & -1.09\% & -1.50\% & -0.93\% & -0.10\% & -0.48\% & -0.75\% \\  
$\sigma$   & 0.0056 & 0.0065 & 0.0094 & 0.0083 & 0.0070 & 0.0044$\dagger$ & 0.0064 & 0.0068 \\ \hline 
NER  & 4.5\% & 0.0\% & 22.7\%$\dagger$ & 0.0\% & \textbf{43.6\%}$\dagger$ & 27.3\%$\dagger$ & 1.8\% & 0.0\% \\
$\Delta F_1$  & -0.85\% & -1.17\% & -0.15\% & -0.73\% &  & -0.08\% & -0.75\% & -0.89\% \\  
$\sigma$   & 0.0073$\dagger$ & 0.0084$\dagger$ & 0.0077$\dagger$ & 0.0081 & 0.0069$\dagger$ & 0.0064$\dagger$ & 0.0081$\dagger$ & 0.0075$\dagger$ \\ \hline 
Entities  & 4.2\% & 7.6\% & 0.8\% & 0.0\% & 6.7\% & \textbf{57.1\%}$\dagger$ & 21.8\% & 1.7\% \\
$\Delta F_1$   & -0.92\% & -0.89\% & -1.50\% & -2.24\% & -0.80\% &  & -0.33\% & -1.13\% \\  
$\sigma$   & 0.0167 & 0.0170 & 0.0178 & 0.0180 & 0.0154 & 0.0148 & 0.0151 & 0.0166 \\ \hline 
Events  & 12.9\% & 4.8\% & 0.0\% & 0.0\% & 0.0\% & \textbf{71.8\%}$\dagger$ & 9.7\% & 0.8\% \\
$\Delta F_1$   & -0.55\% & -0.78\% & -2.77\% & -3.55\% & -2.55\% &  & -0.67\% & -1.36\% \\  
$\sigma$   & 0.0045$\dagger$ & 0.0049$\dagger$ & 0.0098 & 0.0089 & 0.0086 & 0.0060 & 0.0066 & 0.0062 \\ \hline 
\hline
Average  & 17.8\% & 2.5\% & 4.7\% & 0.0\% & 10.1\% & \textbf{57.4\%} & 7.1\% & 0.5\% \\ \hline   
\end{tabular}
\caption{Different randomly sampled configurations were evaluated with each possible pre-trained word embedding. The first number depicts for how many configurations each setting resulted in the best test performance. The second number shows the median difference to the best option for each task. 108 configurations were sampled for POS, 97 for Chunking, 110 for NER, 119 for Entities, and 124 for Events. Statistically significant differences with $p < 0.01$ are marked with $\dagger$.}
\label{table:word_embeddings}
\end{table}

\textbf{Conclusion.} The selection of the pre-trained word embeddings has a large impact on the performance of the system, a much larger impact than many other hyperparameters. On most tasks, the \newcite{Komninos2016} embeddings gave by a far margin the best performance.   

\FloatBarrier
\subsection{Character Representation} \label{sec:char_representation}

We evaluate the approaches of \newcite{Ma2016} using Convolutional Neural Networks (CNN) as well as the approach of \newcite{Lample2016} using LSTM-networks to derive character-based representations.

\reftable{table:char_representation} shows, that character-based representations yield a statistically significant difference only for the POS, the Chunking, and the Events task. For NER and Entities, the difference to not using a character-based representation is not significant ($p > 0.01$).

The difference between the CNN approach by \newcite{Ma2016} and the LSTM approach by \newcite{Lample2016} to derive character-based representations is statistically insignificant. This is quite surprising, as both approaches have fundamentally different properties: The CNN approach from \newcite{Ma2016} takes only trigrams into account. It is also position independent, i.e.\ the network will not be able to distinguish between trigrams at the beginning, in the middle, or at the end of a word, which can be crucial information for some tasks. The BiLSTM approach from \newcite{Lample2016} takes all characters of the word into account. Further, it is position aware, i.e.\ it can distinguish between characters at the start and at the end of the word. Intuitively, one would think that the LSTM approach by Lample et al. would be superior.

\begin{table}[h]
\centering
\begin{tabular}{|c|c|c|c|c|}
\hline
\textbf{Task} & \textbf{\# Configs} & \textbf{No} & \textbf{CNN} & \textbf{LSTM}     \\ \hline
POS & 225 & 4.9\% & \textbf{58.2\%}$\dagger$ & 36.9\% \\
$\Delta Acc.$ &  & -0.90\% &  & -0.05\% \\  
$\sigma$ &  & 0.0201 & 0.0127$\dagger$ & 0.0142$\dagger$ \\ \hline 
Chunking & 241 & 13.3\% & 43.2\%$\dagger$ & \textbf{43.6\%}$\dagger$ \\
$\Delta F_1$ &  & -0.20\% & -0.00\% &  \\  
$\sigma$ &  & 0.0084 & 0.0067$\dagger$ & 0.0065$\dagger$ \\ \hline 
NER & 217 & 27.2\% & \textbf{36.4\%} & 36.4\% \\
$\Delta F_1$ &  & -0.11\% &  & -0.01\% \\  
$\sigma$ &  & 0.0082 & 0.0082 & 0.0080 \\ \hline 
Entities & 228 & 26.8\% & 36.0\% & \textbf{37.3\%} \\
$\Delta F_1$ &  & -0.07\% & 0.00\% &  \\  
$\sigma$ &  & 0.0177 & 0.0165 & 0.0171 \\ \hline 
Events & 219 & 20.5\% & 35.6\%$\dagger$ & \textbf{43.8\%}$\dagger$ \\
$\Delta F_1$ &  & -0.44\% & -0.04\% &  \\  
$\sigma$ &  & 0.0140 & 0.0103$\dagger$ & 0.0096$\dagger$ \\ \hline 
\hline
Average &  & 18.5\% & \textbf{41.9\%} & 39.6\% \\ \hline 
\end{tabular}
\caption{Comparison of not using character-based representations and using CNNs \protect\cite{Ma2016} or LSTMs \protect\cite{Lample2016} to derive character-based representations. The first number depicts for how many configurations each setting resulted in the best test performance. The second number shows the median difference to the best option for each task. Statistically significant differences with $p < 0.01$ are marked with $\dagger$. }
\label{table:char_representation}
\end{table}

Note that we tested both options only with the presented hyperparameters in their respective papers. Each character was mapped to a randomly initialized 30-dimensional embeddings. For the CNN approach, \newcite{Ma2016} used 30 filters and a filter length of 3, which yields a 30 dimensional representation for each word. For the bidirectional LSTM approach, \newcite{Lample2016} used 25 recurrent units, yielding a character-based representation of 50 dimensions for each word. It would of interest if the performance could be improved by selecting different hyperparameters, for example for the CNN approach to not only use character trigrams, but also using shorter and/or longer n-grams.

\textbf{Conclusion.} Character-based representations were in a lot of tested configurations not that helpful and could not improve the performance of the network. The CNN approach by \newcite{Ma2016} and the LSTM approach by \newcite{Lample2016} performed on-par. The CNN approach should be preferred due to the higher computational efficiency.

\FloatBarrier
\subsection{Optimizers} \label{sec:optimizers}
We evaluated different optimizers for the network: Stochastic Gradient Descent (\textit{SGD}), \textit{Adagrad} \cite{Adagrad}, \textit{Adadelta} \cite{Adadelta}, \textit{RMSProp} \cite{RMSProp}, \textit{Adam} \cite{Adam}, and \textit{Nadam} \cite{Nadam}, an Adam variant that incorporates Nesterov momentum \cite{Nesterov:1983}. For SGD, we tuned the learning rate by hand, however, we could observe that it failed in many instances to converge to a minimum. For the other optimizers, we used the recommended settings from the respective papers.

\begin{figure}[ht]
\centering
  \includegraphics[width=\textwidth]{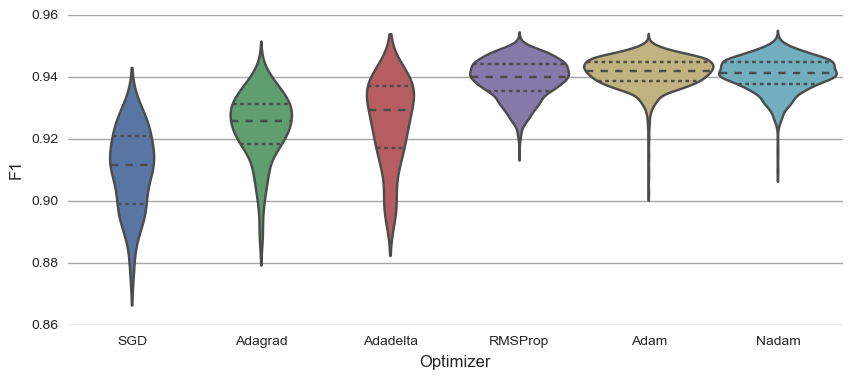}
\caption{Performance on the CoNLL 2000 chunking  shared task for various optimizers.}
\label{fig:optimizer_chunking}
\end{figure}

\begin{figure}[ht]
\centering
  \includegraphics[width=\textwidth]{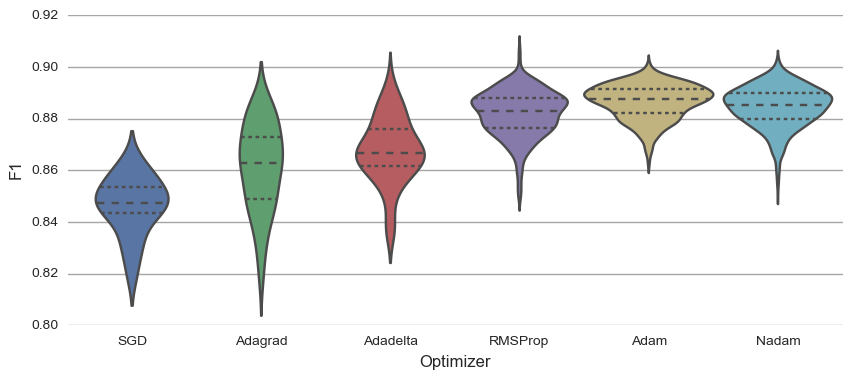}
\caption{Performance on the CoNLL 2003 NER shared task for various optimizers.}
\label{fig:optimizer_ner}
\end{figure}

\reffigure{fig:optimizer_chunking} and \reffigure{fig:optimizer_ner} show the performance for the different choices of optimizers for the Chunking and the NER task, respectively. \reftable{table:optimizers} contains the results for all other tasks.

\begin{table}[h]
\centering
\begin{tabular}{|c|c|c|c|c|c|c|c|c|}
\hline
\textbf{Task} & \textbf{Configs} & \textbf{Adam} & \textbf{Nadam} & \textbf{RMSProp}  & \textbf{Adadelta} & \textbf{Adagrad} & \textbf{SGD}    \\ \hline
POS & 218 & 7.3\% & \textbf{82.6\%}$\dagger$ & 10.1\% & 0.0\% & 0.0\% & 0.0\% \\
$\Delta Acc.$ &  & -0.33\% &  & -0.30\% & -4.08\% & -1.93\% & -18.08\% \\  
$\sigma$ &  & 0.0210$\dagger$ & 0.0172$\dagger$ & 0.0200$\dagger$ & 0.1915 & 0.0463 & 0.2685 \\ 
Median epochs &  & 31 & 20 & 31 & 47 & 40 & 46 \\ \hline 
Chunking & 192 & 17.2\% & \textbf{51.6\%}$\dagger$ & 31.2\%$\dagger$ & 0.0\% & 0.0\% & 0.0\% \\
$\Delta F_1$ &  & -0.11\% &  & -0.04\% & -0.90\% & -1.23\% & -3.74\% \\  
$\sigma$ &  & 0.0090$\dagger$ & 0.0085$\dagger$ & 0.0085$\dagger$ & 0.0125 & 0.0135 & 0.2958 \\ 
Median epochs &  & 16 & 10 & 14 & 37 & 31 & 39 \\ \hline
NER & 207 & 25.1\%$\dagger$ & \textbf{36.7\%}$\dagger$ & 34.8\%$\dagger$ & 1.9\% & 1.4\% & 0.0\% \\
$\Delta F_1$ &  & -0.10\% &  & 0.03\% & -0.77\% & -0.82\% & -5.38\% \\  
$\sigma$ &  & 0.0092$\dagger$ & 0.0091$\dagger$ & 0.0096$\dagger$ & 0.0138 & 0.0139 & 0.1328 \\ 
Median epochs &  & 12 & 9 & 10 & 22 & 19 & 42 \\ \hline 
Entities & 152 & 25.0\%$\dagger$ & \textbf{40.8\%}$\dagger$ & 30.3\%$\dagger$ & 3.9\% & 0.0\% & 0.0\% \\
$\Delta F_1$ &  & -0.14\% &  & -0.09\% & -1.49\% & -1.82\% & -6.00\% \\  
$\sigma$ &  & 0.0167$\dagger$ & 0.0166$\dagger$ & 0.0165$\dagger$ & 0.0244 & 0.0288 & 0.1960 \\ 
Median epochs &  & 13 & 10 & 11 & 32 & 29 & 46 \\ \hline 
Events & 113 & 17.7\%$\dagger$ & \textbf{31.9\%}$\dagger$ & 26.5\%$\dagger$ & 8.0\% & 15.0\% & 0.9\% \\
$\Delta F_1$ &  & -0.10\% &  & -0.05\% & -0.55\% & -0.34\% & -1.68\% \\  
$\sigma$ &  & 0.0129$\dagger$ & 0.0127$\dagger$ & 0.0142$\dagger$ & 0.0142$\dagger$ & 0.0155$\dagger$ & 0.0644 \\ 
Median epochs &  & 8 & 5 & 7 & 12 & 7 & 19 \\ \hline 
\hline
Average &  & 18.5\% & \textbf{48.7\%} & 26.6\% & 2.8\% & 3.3\% & 0.2\% \\ \hline 
\end{tabular}
\caption{Randomly sampled network configurations were evaluated with each optimizer. The first number depicts in how many cases each optimizer produced better results than the others. The second number shows the median difference to the best option for each task. Median epochs depict the median number of train epochs until convergence. Statistically significant differences with $p < 0.01$ are marked with $\dagger$.} 
\label{table:optimizers}
\end{table}

We observe that the variance for RMSProp, Adam, and Nadam is much smaller in comparison to SGD, Adagrad and Adadelta. Here we can conclude that these optimizers are able to achieve a better performance independent of the remaining configuration of the network and/or of the random seed value.

Nadam showed the best performance, yielding the highest score for 48.7\% of the tested configurations. The difference to Adam and RMSProp is often not statistically significant, i.e.\ more experiments would be required to determine which optimizer yields the best performance.

We measured the time an optimizer required to converge. The time per epoch was for all optimizers similar, however, we observed large difference in terms of number of epochs until convergence. Nadam converged the fastest, i.e.\ requiring only a small number of trainings epoch to achieve good performance. Adadelta, Adagrad and SGD required the longest to converge to a minimum.

\newcite{Adam} recommend for the Adam optimizer certain default parameters. However, we did some manual tuning for our tasks and increased the learning rate for Adam to $0.01$ for the first three epochs, set it to $0.005$ for the consecutive three epochs and then setting it back to its default value of $0.001$. The result is depicted in \reftable{table:adam-optimized}. Not only does this lead to a faster convergence, it also leads to a better performance for the POS tagging and the Chunking task. It would be of interest to evaluate the other hyperparameters of Adam and Nadam to see their impact on the performance.

\begin{table}[h]
\centering
\begin{tabular}{|c|c|c|c|}
\hline
\textbf{Task} & \textbf{Configs} & \textbf{Adam} & \textbf{Adam (increased LR)}  \\ \hline
POS & 226 & 35.0\% & \textbf{65.0\%} \\
$\Delta Acc.$ &  & (-0.15\%) &  \\ 
Median train epochs &  & (31) & (22) \\ \hline 
Chunking & 203 & 31.0\% & \textbf{69.0\%} \\
$\Delta F_1$ &  & (-0.17\%) &  \\ 
Median train epochs &  & (15) & (7) \\ \hline 
NER & 224 & 46.9\% & \textbf{53.1\%} \\
$\Delta F_1$ &  & (-0.08\%) &  \\ 
Median train epochs &  & (12) & (7) \\ \hline 
Entities & 197 & \textbf{56.3\%} & 43.7\% \\
$\Delta F_1$ &  &  & (-0.15\%) \\ 
Median train epochs &  & (16) & (9) \\ \hline 
Events & 210 & \textbf{50.5\%} & 49.5\% \\
$\Delta F_1$ &  &  & (-0.00\%) \\ \
Median train epochs &  & (8) & (5) \\ \hline 
\hline
Average &  & 43.9\% & \textbf{56.1\%} \\ \hline  
\end{tabular}
\caption{Comparison of Adam to Adam with increase learning rate (LR). The first number depicts in how many cases each optimizer produced better results than the others. The second number shows the median difference to the best option for each task. Median epochs depict the median number of train epochs until convergence.}
\label{table:adam-optimized}
\end{table}

\textbf{Conclusion.} Adam, Nadam, and RMSProp produced more stable and better results than SGD, Adagrad or Adadelta.  In our experiments, Nadam (Adam with Nesterov momentum) was on average the best optimizer. RMSProp produced test scores on average up to 0.30 percentage points below of Adam or Nadam. Nadam had the best convergence time. Adapting the learning rate for Adam can further improve the performance as well as the convergence time.

\FloatBarrier
\subsection{Gradient Clipping and Normalization} \label{sec:gradient_clipping}
Two common strategies to deal with the exploding gradient problem are \textit{gradient clipping} \cite{Mikolov2012} and \textit{gradient normalization} \cite{Pascanu2013}. Gradient clipping involves clipping the gradient's components element-wise if they exceeds a defined threshold. Gradient normalization has a better theoretical justification and rescales the gradient whenever the norm goes over a threshold.

The results for gradient clipping are depicted in \reftable{table:clipvalue}. For the evaluated threshold, we could not observe any statistically significant improvement for the five tasks.

Gradient normalization has a better theoretical justification \cite{Pascanu2013} and we can clearly observe that it leads to a better performance as depicted in \reftable{table:clipnorm}. The concrete threshold value for the gradient normalization is of lower importance, as long as it is not too small or too large. A threshold value of 1 performed the best in most cases.

\begin{table}[h]
\centering
\begin{tabular}{|c|c|c|c|c|c|c|}
\cline{3-7}
\multicolumn{2}{c|}{} & \multicolumn{5}{c|}{\bfseries Clipping threshold } \\ \hline
\textbf{Task} & \textbf{\# Configs} & \textbf{None} & \textbf{1} & \textbf{3} & \textbf{5} & \textbf{10}  \\ \hline
POS & 106 & \textbf{24.5\%} & 21.7\% & 20.8\% & 17.9\% & 15.1\% \\
$\Delta Acc.$ &  &  & -0.00\% & -0.02\% & -0.02\% & -0.02\% \\  
$\sigma$ &  & 0.2563 & 0.2649 & 0.2407 & 0.2809 & 0.2715 \\ \hline 
Chunking & 109 & 24.8\% & \textbf{26.6\%} & 15.6\% & 13.8\% & 19.3\% \\
$\Delta F_1$ &  & -0.05\% &  & -0.02\% & -0.03\% & -0.03\% \\  
$\sigma$ &  & 0.1068 & 0.0101 & 0.0765 & 0.1160 & 0.0111 \\ \hline 
NER & 84 & 16.7\% & \textbf{25.0\%} & 22.6\% & 17.9\% & 17.9\% \\
$\Delta F_1$ &  & -0.04\% &  & -0.00\% & -0.03\% & 0.02\% \\  
$\sigma$ &  & 0.0110 & 0.0110 & 0.0104 & 0.0111 & 0.0114 \\ \hline 
Entities & 85 & 18.8\% & 16.5\% & 21.2\% & \textbf{22.4\%} & 21.2\% \\
$\Delta F_1$ &  & -0.10\% & -0.07\% & -0.04\% &  & -0.07\% \\  
$\sigma$ &  & 0.0176 & 0.0167 & 0.0188 & 0.0190 & 0.0203 \\ \hline 
Events & 99 & 21.2\% & 17.2\% & 16.2\% & \textbf{28.3\%} & 17.2\% \\
$\Delta F_1$ &  & -0.02\% & -0.14\% & -0.01\% &  & -0.05\% \\  
$\sigma$ &  & 0.0156 & 0.0161 & 0.0167 & 0.0158 & 0.0153 \\ \hline 
\hline
Average &  & 21.2\% & \textbf{21.4\%} & 19.3\% & 20.0\% & 18.1\% \\ \hline 
\end{tabular}
\caption{Element-wise clipping of gradient values to a certain threshold, as described by \protect\newcite{Mikolov2012}. The tested thresholds 1, 3, 5 and 10 did not lead to any improvement compared to not clipping the gradient. Statistically significant differences with $p < 0.01$ are marked with $\dagger$.}
\label{table:clipvalue}
\end{table}

\begin{table}[h]
\centering
\begin{tabular}{|c|c|c|c|c|c|c|}
\cline{3-7}
\multicolumn{2}{c|}{} & \multicolumn{5}{c|}{\bfseries Normalization threshold } \\ \hline
\textbf{Task} & \textbf{\# Configs} & \textbf{None} & \textbf{1} & \textbf{3} & \textbf{5} & \textbf{10}  \\ \hline
POS & 106 & 3.8\% & \textbf{40.6\%}$\dagger$ & 24.5\%$\dagger$ & 15.1\%$\dagger$ & 16.0\% \\
$\Delta Acc.$ &  & -0.82\% &  & -0.05\% & -0.05\% & -0.08\% \\  
$\sigma$ &  & 0.2563 & 0.0254$\dagger$ & 0.0245$\dagger$ & 0.0253$\dagger$ & 0.0254$\dagger$ \\ \hline 
Chunking & 109 & 6.4\% & \textbf{27.5\%}$\dagger$ & 24.8\%$\dagger$ & 22.0\%$\dagger$ & 19.3\%$\dagger$ \\
$\Delta F_1$ &  & -0.29\% &  & -0.00\% & -0.04\% & -0.04\% \\  
$\sigma$ &  & 0.1068 & 0.0087 & 0.0087 & 0.0088 & 0.0086 \\ \hline 
NER & 84 & 7.1\% & \textbf{32.1\%}$\dagger$ & 20.2\%$\dagger$ & 23.8\%$\dagger$ & 16.7\%$\dagger$ \\
$\Delta F_1$ &  & -0.44\% &  & -0.05\% & -0.10\% & -0.14\% \\  
$\sigma$ &  & 0.0110 & 0.0101 & 0.0101 & 0.0100 & 0.0112 \\ \hline 
Entities & 87 & 6.9\% & 21.8\%$\dagger$ & 23.0\%$\dagger$ & \textbf{27.6\%}$\dagger$ & 20.7\%$\dagger$ \\
$\Delta F_1$ &  & -0.58\% & -0.02\% & 0.09\% &  & -0.01\% \\  
$\sigma$ &  & 0.0831 & 0.0168 & 0.0158 & 0.0172 & 0.0163 \\ \hline 
Events & 106 & 3.8\% & \textbf{30.2\%}$\dagger$ & 26.4\%$\dagger$ & 21.7\%$\dagger$ & 17.9\%$\dagger$ \\
$\Delta F_1$ &  & -0.59\% &  & -0.03\% & -0.09\% & -0.21\% \\  
$\sigma$ &  & 0.0157 & 0.0143 & 0.0137 & 0.0147 & 0.0141 \\ \hline 
\hline
Average &  & 5.6\% & \textbf{30.5\%} & 23.8\% & 22.0\% & 18.1\% \\ \hline 
\end{tabular}
\caption{Normalizing the gradient to $\hat g = (\tau/||g||_2) g$  if the norm $||g||_2$ exceeds a threshold $\tau$ \protect\cite{Pascanu2013}. We see a clear performance increase in comparison to not normalizing the gradient. The threshold value $\tau$ is of minor importance, with $\tau=1$ usually performing the best. Statistically significant differences with $p < 0.01$ are marked with $\dagger$.}
\label{table:clipnorm}
\end{table}

\textbf{Conclusion.} Gradient clipping does not improve the performance. Gradient normalization as described by \newcite{Pascanu2013} improves significantly the performance with an observed average improvement between 0.45 and 0.82 percentage points. The threshold $\tau$ is of minor importance, with $\tau=1$ giving usually the best results.  

\FloatBarrier
\subsection{Tagging Schemes} \label{sec:tagging_schemes}

\begin{figure}[ht]
\centering
  \includegraphics[width=\textwidth]{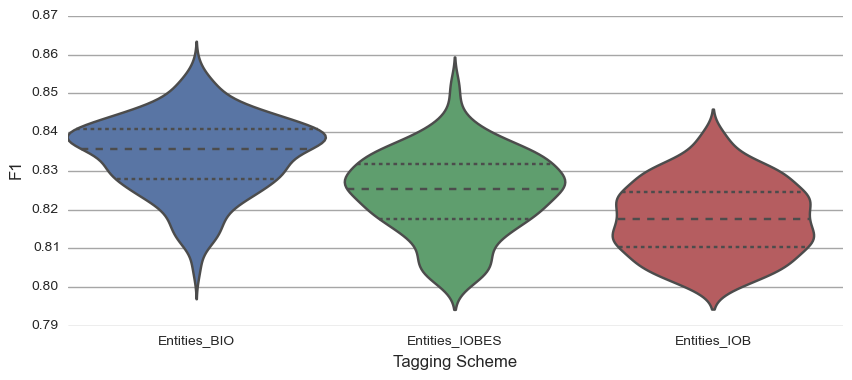}
\caption{Performance on the ACE 2005 entities dataset for various tagging schemes.}
\label{fig:tagging_entities}
\end{figure}

\reffigure{fig:tagging_entities} depicts the violin plot for different tagging schemes for the ACE 2005 entities recognition task. \reftable{table:tagging} summarizes the results for all other tasks. The IOB tagging scheme performs poorly on all tasks. The BIO and IOBES tagging schemes perform on-par, except for the Entities dataset, here we observe a much better performance for the BIO scheme. 

\begin{table}[h]
\centering
\begin{tabular}{|c|c|c|c|c|c|}
\hline
\textbf{Task} &  \textbf{\# Configs} & \textbf{BIO} & \textbf{IOB} & \textbf{IOBES}  \\ \hline
Chunking & 106 & 38.7\%$\dagger$ & 4.7\% & \textbf{56.6\%}$\dagger$ \\
$\Delta F_1$ &  & -0.07\% & -0.34\% &  \\  
$\sigma$ &  & 0.0052 & 0.0061 & 0.0048 \\ \hline 
NER & 98 & 40.8\%$\dagger$ & 7.1\% & \textbf{52.0\%}$\dagger$ \\
$\Delta F_1$ &  & -0.09\% & -0.46\% &  \\  
$\sigma$ &  & 0.0074 & 0.0084 & 0.0066 \\ \hline 
Entities & 106 & \textbf{88.7\%}$\dagger$ & 0.0\% & 11.3\% \\
$\Delta F_1$ &  &  & -1.90\% & -1.01\% \\  
$\sigma$ &  & 0.0108$\dagger$ & 0.0162 & 0.0142$\dagger$ \\ \hline 
Events & 107 & \textbf{47.7\%}$\dagger$ & 9.3\% & 43.0\%$\dagger$ \\
$\Delta F_1$ &  &  & -0.34\% & -0.05\% \\  
$\sigma$ &  & 0.0038 & 0.0039 & 0.0044 \\ \hline \hline
Average &  & \textbf{54.0\%} & 5.3\% & 40.7\% \\ \hline 
\end{tabular}
\caption{Network configurations were sampled randomly and were evaluated with each tagging scheme. The first number in the cell depicts in how many cases each tagging scheme produced better results than the others. The second number shows the median difference to the best option for each task. Statistically significant differences with $p < 0.01$ are marked with $\dagger$.}

\label{table:tagging}

\end{table}

\textbf{Conclusion.} The IOB tagging scheme produced by far the worst results. The BIO and IOBES tagging schemes were producing similar results. We would recommend using the BIO tagging scheme as the generated overhead is smaller than for the IOBES scheme.

\FloatBarrier
\subsection{Classifier - Softmax vs. CRF}\label{sec:classifiers}
\FloatBarrier

\begin{figure}[h]
\centering
\begin{subfigure}{.32\textwidth}
  \centering
  \includegraphics[width=\linewidth]{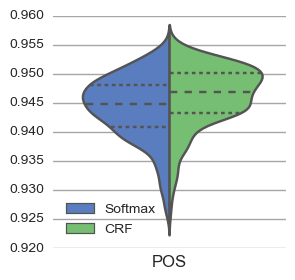}
\end{subfigure}
\begin{subfigure}{.32\textwidth}
  \centering
  \includegraphics[width=\linewidth]{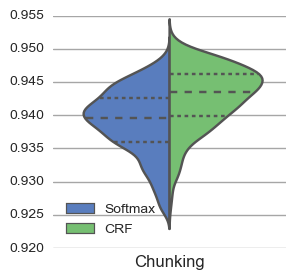}
\end{subfigure}
\begin{subfigure}{.32\textwidth}
  \centering
  \includegraphics[width=\linewidth]{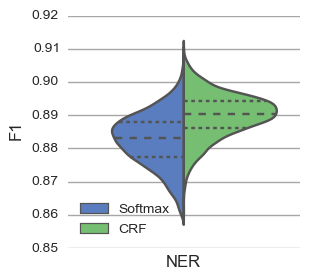}
\end{subfigure}
\begin{subfigure}{.32\textwidth}
  \centering
  \includegraphics[width=\linewidth]{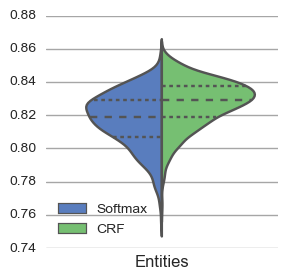}
\end{subfigure}
\begin{subfigure}{.32\textwidth}
  \centering
  \includegraphics[width=\linewidth]{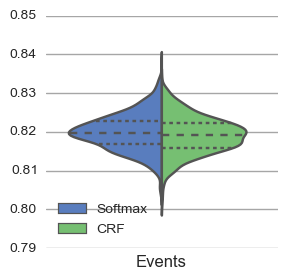}
\end{subfigure}
\caption{Softmax versus CRF classifier for the examined benchmark tasks. The plot shows the accuracy for the POS tagging task and the $F_1$-score for the other tasks. The violin plots show clearly better results for the CRF classifier all except the TempEval3 events task. }
\label{fig:classifier_comparison}
\end{figure}

\reffigure{fig:classifier_comparison} depicts the difference between a Softmax classifier as final layer versus optimizing the complete label sequence for the whole sentence using a Conditional Random Field (CRF). The violin plots show a clear preference for a CRF classifier as final layer for all except the TempEval 3 events dataset. \reftable{table:classifier} compares the two options when all other hyperparameters are kept the same. It confirms the impression that CRF leads to superior results in most cases, except for the event detection task. The improvement by using a CRF classifier instead of a softmax classifier lies between 0.19 percentage points and 0.85 percentage points for the evaluated tasks.

When using the BIO- or IOBES tagging scheme, we observe that a softmax classifier produces a high number of invalid tags, e.g.\ an \texttt{I-} tag starts without a previous \texttt{B-} tag. For example, a system with a single BiLSTM-layer and a softmax classifier produced on the test set for around 1.5\% - 2.0\% of the named entities an invalid BIO tagging. Increasing the number of BiLSTM reduces the number of invalid tags: A system with three BiLSTM-layers produced invalid tags for around 0.3\% - 0.8\% of the named entities. We evaluated two strategies for correcting invalid tags: Either setting them to \texttt{O} or setting them to the \texttt{B-} tag to start a new segment. However, the difference between these two strategies is negligible. The CRF classifier, on the other hand, produces in most cases no invalid tags and only in rare case are one or two named entities wrongly tagged. 

\begin{figure}[ht]
\centering
  \includegraphics[width=\textwidth]{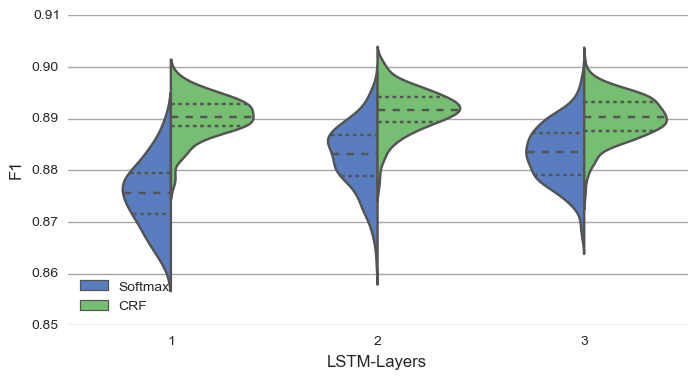}
\caption{Difference between Softmax and CRF classifier for different numbers of BiLSTM-layers for the CoNLL 2003 NER dataset.}
\label{fig:decoder_ner_lstm_layers}
\end{figure}

\reffigure{fig:decoder_ner_lstm_layers} depicts the difference between a Softmax and a CRF classifier for different numbers of stacked BiLSTM-layers for the NER task. The violin plot shows, that a CRF classifier brings the largest improvement for shallow BiLSTM-networks. With an increasing number of BiLSTM-layers, the difference decreases. However, for depth 3 we still observe in the plot a significant difference between the two classifiers. The figure also shows that when using a softmax classifier, more BiLSTM-layers are beneficial, however, when using a CRF classifier, the difference between 1, 2, or 3 BiLSTM layers is much smaller. This effect is further studied in \refsec{sec:num_lstm_layers}.

\begin{table}[h]
\centering
\begin{tabular}{|c|c|c|c|c|}
\hline
\textbf{Task} & \textbf{\# Configs} & \textbf{Softmax} & \textbf{CRF}   \\ \hline
POS & 114 & 19.3\% & \textbf{80.7\%}$\dagger$ \\
$\Delta Acc.$ &  & -0.19\% &  \\  
$\sigma$ &  & 0.0149 & 0.0132 \\ \hline 
Chunking & 230 & 4.8\% & \textbf{95.2\%}$\dagger$ \\
$\Delta F_1$ &  & -0.38\% &  \\  
$\sigma$ &  & 0.0058 & 0.0051 \\ \hline 
NER & 235 & 9.4\% & \textbf{90.6\%}$\dagger$ \\
$\Delta F_1$ &  & -0.67\% &  \\  
$\sigma$ &  & 0.0081 & 0.0060$\dagger$ \\ \hline 
Entities & 214 & 13.1\% & \textbf{86.9\%}$\dagger$ \\
$\Delta F_1$ &  & -0.85\% &  \\  
$\sigma$ &  & 0.0157 & 0.0140 \\ \hline 
Events & 203 & \textbf{61.6\%}$\dagger$ & 38.4\% \\
$\Delta F_1$ &  &  & -0.15\% \\  
$\sigma$ &  & 0.0052 & 0.0057 \\ \hline 
\hline
Average &  & 21.6\% & \textbf{78.4\%} \\ \hline 
\end{tabular}

\caption{Network configurations were sampled randomly and each was evaluated with each classifier as a last layer. The first number in a cell depicts in how many cases each classifier produced better results than the others. The second number shows the median difference to the best option for each task. Statistically significant differences with $p < 0.01$ are marked with $\dagger$.}

\label{table:classifier}

\end{table}

For the TempEval 3 event dataset, Softmax is slightly better than a CRF classifier. This is due to the distribution of the labels: The \texttt{I-} tag appears only three times in the whole corpus and not once in the training data. Each event in the training data is therefore composed of only a single token. Hence, except for the tree events in the test dataset, there are no dependencies between the tags and a CRF classifier does not bring any improvement. 

\textbf{Conclusion.} In case there are dependencies between the labels CRF usually outperforms Softmax as the last layer of the neural network. This outperformance is also true for stacked BiLSTM-layers. In case there are no or only negligible dependencies between labels in a sentence, Softmax performs better.

\FloatBarrier
\subsection{Dropout}\label{sec:dropout}

We compare \textit{no dropout}, \textit{naive dropout}, and \textit{variational dropout} \cite{Gal2015}. Naive dropout applies a new dropout mask at every time step of the LSTM-layer. Variational dropout applies the same dropout mask for all time steps in the same sentence. Further, it applies dropout to the recurrent units. We evaluate the dropout rates $\{0.05, 0.1, 0.25, 0.5\}$.

\reftable{table:dropout} depicts the results for the different dropout schemes. We observe, that variational dropout results in most cases to the best performance. The median difference to not using dropout can be as high as $\Delta F_1=-1.98\%$ for the Entities task and $\Delta F_1=-1.32\%$ in comparison to naive dropout. For all tasks it yielded the lowest standard deviation, indicating that variational dropout makes the network more robust in terms of the selected hyperparameters and / or the random seed value.

\begin{table}[h]
\centering
\begin{tabular}{|c|c|c|c|c|c|}
\hline
\textbf{Task} & \textbf{\# Configs} & \textbf{No} & \textbf{Naive} & \textbf{Variational}  \\ \hline
POS & 77 & 6.5\% & 19.5\% & \textbf{74.0\%}$\dagger$ \\
$\Delta Acc.$ &  & -0.27\% & -0.18\% &  \\  
$\sigma$ &  & 0.0083 & 0.0108 & 0.0076 \\ \hline 
Chunking & 91 & 0.0\% & 4.4\% & \textbf{95.6\%}$\dagger$ \\
$\Delta F_1$ &  & -0.88\% & -0.53\% &  \\  
$\sigma$ &  & 0.0055 & 0.0053 & 0.0037$\dagger$ \\ \hline 
NER & 127 & 3.9\% & 7.9\% & \textbf{88.2\%}$\dagger$ \\
$\Delta F_1$ &  & -0.79\% & -0.54\% &  \\  
$\sigma$ &  & 0.0077 & 0.0075 & 0.0059 \\ \hline 
Entities & 90 & 2.2\% & 6.7\% & \textbf{91.1\%}$\dagger$ \\
$\Delta F_1$ &  & -1.98\% & -1.32\% &  \\  
$\sigma$ &  & 0.0159 & 0.0155$\dagger$ & 0.0119$\dagger$ \\ \hline 
Events & 97 & 15.5\% & 17.5\% & \textbf{67.0\%}$\dagger$ \\
$\Delta F_1$ &  & -0.47\% & -0.28\% &  \\  
$\sigma$ &  & 0.0054 & 0.0051 & 0.0038 \\ \hline 
\hline
Average &  & 5.6\% & 11.2\% & \textbf{83.2\%} \\ \hline 
\end{tabular}
\caption{Network configurations were sampled randomly and each was evaluated with each dropout scheme. The first number in a cell depicts in how many cases each dropout scheme produced better results than the others. The second number shows the median difference to the best option for each task. Statistically significant differences with $p < 0.01$ are marked with $\dagger$.}

\label{table:dropout}

\end{table}

\reftable{table:variational_dropout} evaluates variational dropout which is applied either to the output or to the recurrent units. We observe that variational dropout should be applied to both units.

\begin{table}[h]
\centering
\begin{tabular}{|c|c|c|c|c|c|}
\hline
\textbf{Task} & \textbf{\# Configs} & \textbf{Output} & \textbf{Recurrent} & \textbf{Both}  \\ \hline
POS & 95 & 3.2\% & 37.9\%$\dagger$ & \textbf{58.9\%}$\dagger$ \\
$\Delta Acc.$ &  & -0.29\% & -0.05\% &  \\  
$\sigma$ &  & 0.0139 & 0.0119 & 0.0171 \\ \hline 
Chunking & 163 & 14.1\% & 18.4\% & \textbf{67.5\%}$\dagger$ \\
$\Delta F_1$ &  & -0.32\% & -0.25\% &  \\  
$\sigma$ &  & 0.0050 & 0.0053 & 0.0050 \\ \hline 
NER & 144 & 9.7\% & 22.9\% & \textbf{67.4\%}$\dagger$ \\
$\Delta F_1$ &  & -0.42\% & -0.34\% &  \\  
$\sigma$ &  & 0.0074 & 0.0075 & 0.0063 \\ \hline 
Entities & 144 & 9.7\% & 25.0\% & \textbf{65.3\%}$\dagger$ \\
$\Delta F_1$ &  & -0.82\% & -0.64\% &  \\  
$\sigma$ &  & 0.0149 & 0.0142 & 0.0113 \\ \hline 
Events & 158 & 29.7\% & 15.8\% & \textbf{54.4\%}$\dagger$ \\
$\Delta F_1$ &  & -0.15\% & -0.33\% &  \\  
$\sigma$ &  & 0.0048 & 0.0042$\dagger$ & 0.0034$\dagger$ \\ \hline 
\hline
Average &  & 13.3\% & 24.0\% & \textbf{62.7\%} \\ \hline 
\end{tabular}
\caption{Network configurations were sampled randomly and each was evaluated with different dropout rates for variational dropout. The first number in a cell depicts in how many cases each variational dropout scheme produced better results than the others. The second number shows the median difference to the best option for each task. Statistically significant differences with $p < 0.01$ are marked with $\dagger$.}
\label{table:variational_dropout}
\end{table}

\textbf{Conclusion.} Variational dropout was on all tasks superior to no-dropout or naive dropout. Applying dropout along the vertical as well as the recurrent dimension achieved on all benchmark tasks the best result.

\FloatBarrier
\subsection{Going deeper - Number of LSTM-Layers} \label{sec:num_lstm_layers}
We sampled hyperparameters and selected randomly a value $60 \leq u \leq 300$ that is divisible by 2 and 3. We then trained one network with a single BiLSTM layer, one network with two BiLSTM layers, and one and with three BiLSTM layers. The recurrent units per LSTM was set to $u / \#layers$, for example, with $u=150$ and two layers, we have two BiLSTM-layers, each of the four LSTMs has 75 recurrent units.

The result is depicted in \reftable{table:num_lstm_layers}. For POS-tagging, one and two layers performed the best, for Chunking and NER, two or three layers performed the best, for the Entities tasks two layers performed best and for the Events task, there is no large enough difference between these three options. In conclusion two BiLSTM layers appears a robust rule of thumb for sequence tagging.

\begin{table}[h]
  \centering
  \begin{tabular}{|c|c|c|c|c|c|}
\cline{3-5}
\multicolumn{2}{c|}{} & \multicolumn{3}{c|}{\bfseries Num. LSTM-Layers } \\ \hline
  \textbf{Task} & \textbf{\# Configs} & \textbf{1} & \textbf{2} & \textbf{3}  \\ \hline
POS & 64 & \textbf{51.6\%}$\dagger$ & 46.9\%$\dagger$ & 1.6\% \\
$\Delta Acc.$ &  &  & -0.02\% & -0.73\% \\  
$\sigma$ &  & 0.0038 & 0.0034 & 0.0154 \\ \hline 
Chunking & 92 & 10.9\% & \textbf{52.2\%}$\dagger$ & 37.0\%$\dagger$ \\
$\Delta F_1$ &  & -0.29\% &  & -0.11\% \\  
$\sigma$ &  & 0.0059 & 0.0045 & 0.0042 \\ \hline 
NER & 84 & 7.1\% & \textbf{54.8\%}$\dagger$ & 38.1\%$\dagger$ \\
$\Delta F_1$ &  & -0.53\% &  & -0.20\% \\  
$\sigma$ &  & 0.0105 & 0.0082 & 0.0079 \\ \hline 
Entities & 75 & 21.3\% & \textbf{52.0\%}$\dagger$ & 26.7\% \\
$\Delta F_1$ &  & -0.72\% &  & -0.34\% \\  
$\sigma$ &  & 0.0152 & 0.0128 & 0.0135 \\ \hline 
Events & 73 & 30.1\% & \textbf{47.9\%} & 21.9\% \\
$\Delta F_1$ &  & -0.11\% &  & -0.20\% \\  
$\sigma$ &  & 0.0050 & 0.0041 & 0.0044 \\ \hline 
\hline
Average &  & 24.2\% & \textbf{50.8\%} & 25.0\% \\ \hline 
  \end{tabular}
\caption{Network configurations were sampled randomly and each was evaluated with each possible number of stacked BiLSTM-layers. The number of recurrent units is the same for all evaluated depths. The first number in a cell depicts in how many cases each depth produced better results than the others. The second number shows the median difference to the best option for each task. Statistically significant differences with $p < 0.01$ are marked with $\dagger$.}

\label{table:num_lstm_layers}

\end{table}

\textbf{Conclusion.} Except for the reduced POS tagging task, two BiLSTM-layers produced the best results.

\FloatBarrier
\subsection{Going wider - Number of Recurrent Units} \label{sec:num_recurrent_units}
Finding the optimal number of recurrent units is due to the large number of possibilities not straight forward. If the number of units is too small, the network will not be able to store all necessary information to solve the task optimally. If the number is too big, the network will overfit on the trainings data and we will observe declining test performance. But as the performance depends on many other factors, and as shown in \reftable{table:random_init} also heavily on the seed value of the random number generator, simply testing different recurrent unit sizes and choosing the one with the highest performance will result in wrong conclusions. 

To still answer the question of the optimal number of recurrent units as well as how large the impact of this hyperparameter is, we decided to use a method that, due to the large number of runs, is robust to noise from other sources. We decided to compute a polynomial regression between the number of recurrent units and the test performance. We chose a polynomial of degree 2, as we expect that the network will have peak performance at some number of recurrent units and performance will decrease if we choose a smaller value (underfitting) or a larger value (overfitting).

For the polynomial regression, we search for the parameters $a$, $b$, and $c$ that minimize the squared error:
$$ E = \sum_{j=0}^k | p(x_j) - y_j|^2$$

with $p(x) = ax^2+bx+c$,  $x_j$ the average number of recurrent units per LSTM-network, and $y_j$ the performance on the test set for the samples $j=0,...k$.

\begin{figure}[ht]
\centering
  \includegraphics[width=0.8\textwidth]{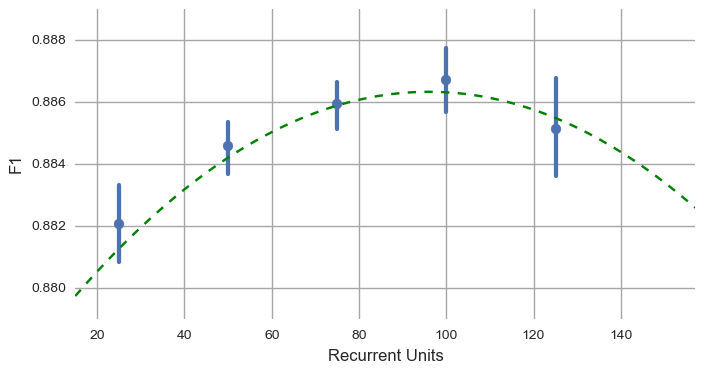}
\caption{Polynomial regression (green, dashed line) for the NER dataset with two stacked BiLSTM-layers.  The blue bars are the median and the 95\% confidence interval of the median for evenly-spaced bins at different center positions.}
\label{fig:num_units_ner_2}
\end{figure}

\reffigure{fig:num_units_ner_2} illustrates the polynomial regression for the NER task with a two stacked  BiLSTM-network. Note, the depicted number of recurrent units is the number of units per LSTM-network. As bidirectional LSTM-networks are used, there are in total 4 LSTM-networks, hence, the total number of recurrent units in the network is 4 times higher than depicted on the x-axis.

We use the polynomial $p(x)$ to find analytically the maximum $x_\text{opt}$, i.e.\ the number of recurrent units that give on average the best test performance. We also use the polynomial to determine how large the impact of this hyperparameter is. A flat polynomial (small $a$ value) is rather robust against changes in this parameters. Selecting a non-optimal number is less important and it would not be worthwhile to optimize this hyperparameter heavily. A steep polynomial (large $a$ value) is more sensitive, a slightly too small or too large number of recurrent units changes the performance significantly. To make this intuitive understandable, we computed $\gamma_{25} = p(x_\text{opt} \pm 25) - p(x_\text{opt})$, which depicts how much the test performance will decrease if we choose the number of recurrent units either 25 units too small or too large.  \reftable{table:num_lstm_units} summarizes our results.

As the table shows, the number of optimal recurrent units (per direction) depends on the task and the number of stacked BiLSTM units. The optimal values lay at around 100. However, as the value $\gamma_{25}$ reveals, this hyperparameter has a rather small impact on the results. Adding or removing 25 recurrent units from a two stacked BiLSTM-network changes the performance by only roughly 0.01\% up to 0.06\%.

\begin{table}[h]
\centering
\begin{tabular}{|c|c|c|c|c|}
\cline{2-4}
\multicolumn{1}{c|}{} & \multicolumn{3}{c|}{\bfseries LSTM-Layers } \\ \hline
\textbf{Task} & \textbf{1} & \textbf{2} & \textbf{3}  \\ \hline
POS & 157 & 63 & 103 \\
$\gamma_{25}$ & -0.03\% & -0.01\% & -0.15\% \\ \hline
Chunking & 174 & 106 & 115 \\
$\gamma_{25}$ & -0.01\% & -0.05\% & -0.03\% \\ \hline
NER & 115 & 96 & 92 \\
$\gamma_{25}$ & -0.01\% & -0.06\% & -0.07\% \\ \hline
Entities & 192 & 175 & 115 \\
$\gamma_{25}$ & -0.04\% & -0.04\% & -0.10\% \\ \hline
Events & 126 & 56 & - \\
$\gamma_{25}$ & -0.01\% & -0.03\% & - \\ \hline
\end{tabular}
\caption{The first number in each cell is the optimal number of recurrent units $x_\text{opt}$ per LSTM-network. The second number shows the value $\gamma_{25} = p(x_\text{opt} \pm 25) - p(x_\text{opt})$, i.e.\ when changing the number of recurrent units by 25, how much does the test performance change. For the Events dataset with 3 stacked BiLSTM-layers, the optimal number was not in the tested range and hence was not found by the polynomial regression approach.  }
\label{table:num_lstm_units}
\end{table}

\textbf{Conclusion.} The number of recurrent units, as long as it is not far too large or far too small, has only a minor effect on the results. A value of about 100 for each LSTM-network appears to be a good rule of thumb for the tested tasks.

\FloatBarrier
\subsection{Mini-Batch Size} \label{sec:mini-batch-size}
We evaluated the mini-batch sizes 1, 8, 16, 32, and 64. The results are depicted in \reftable{table:miniBatchSize}. It appears that for tasks with small training sets a smaller mini-batch size of 1 up to 16 is a good choice. For larger training sets appears 8 - 32 a good choice. The largest difference was seen for the ACE 2005 Entities dataset, where changing the mini-batch size to 1 decreased the median performance by 2.83 percentage points compared to a mini-batch size of 32.

\begin{table}[h]
\centering
\begin{tabular}{|c|c|c|c|c|c|c|}
\cline{3-7}
\multicolumn{2}{c|}{} & \multicolumn{5}{c|}{\bfseries Mini-Batch Size } \\ \hline
\textbf{Task} & \textbf{\# Configs} & \textbf{1} & \textbf{8} & \textbf{16} & \textbf{32} & \textbf{64}  \\ \hline
POS & 102 & \textbf{51.0\%}$\dagger$ & 27.5\%$\dagger$ & 9.8\% & 5.9\% & 5.9\% \\
$\Delta Acc.$ &  &  & -0.07\% & -0.16\% & -0.26\% & -0.20\% \\  
$\sigma$ &  & 0.0169 & 0.0164 & 0.0175 & 0.0183 & 0.0179 \\ \hline 
Chunking & 94 & 9.6\% & \textbf{40.4\%}$\dagger$ & 27.7\%$\dagger$ & 16.0\% & 6.4\% \\
$\Delta F_1$ &  & -0.56\% &  & -0.05\% & -0.10\% & -0.22\% \\  
$\sigma$ &  & 0.0556 & 0.0089$\dagger$ & 0.0092$\dagger$ & 0.0092$\dagger$ & 0.0091$\dagger$ \\ \hline 
NER & 106 & 5.7\% & 16.0\%$\dagger$ & 22.6\%$\dagger$ & \textbf{30.2\%}$\dagger$ & 25.5\%$\dagger$ \\
$\Delta F_1$ &  & -1.11\% & -0.27\% & -0.18\% &  & -0.10\% \\  
$\sigma$ &  & 0.0686 & 0.0120$\dagger$ & 0.0096$\dagger$ & 0.0090$\dagger$ & 0.0099$\dagger$ \\ \hline 
Entities & 107 & 2.8\% & 23.4\%$\dagger$ & 25.2\%$\dagger$ & \textbf{33.6\%}$\dagger$ & 15.0\% \\
$\Delta F_1$ &  & -2.83\% & -0.21\% & -0.07\% &  & -0.31\% \\  
$\sigma$ &  & 0.0793 & 0.0168$\dagger$ & 0.0157$\dagger$ & 0.0159$\dagger$ & 0.0170$\dagger$ \\ \hline 
Events & 91 & \textbf{33.0\%}$\dagger$ & 25.3\%$\dagger$ & 27.5\%$\dagger$ & 6.6\%$\dagger$ & 7.7\% \\
$\Delta F_1$ &  &  & 0.00\% & 0.09\% & -0.32\% & -0.46\% \\  
$\sigma$ &  & 0.0253 & 0.0144 & 0.0133 & 0.0130 & 0.0139 \\ \hline 
\hline
Average &  & 20.4\% & \textbf{26.5\%} & 22.6\% & 18.5\% & 12.1\% \\ \hline 
\end{tabular}
\caption{Network configurations were sampled randomly and each was evaluated with different mini-batch sizes. The first number in a cell depicts in how many cases each mini-batch size  produced better results than the others. The second number shows the median difference to the best option for each task. Statistically significant differences with $p < 0.01$ are marked with $\dagger$.}
\label{table:miniBatchSize}
\end{table}

\textbf{Conclusion.} For tasks with small training sets appears a mini-batch size of 8 a robust selection. For tasks with larger training sets appears a mini-batch size of 32 a robust selection.

\FloatBarrier
\subsection{Theano vs. Tensorflow} \label{sec:theano_vs_tensorflow}
Keras offers the option to choose either Theano\footnote{Theano version 0.8.2} or Tensorflow\footnote{Tensorflow version 0.12.1} as backend. Due to slightly
different implementations of the mathematical operations and numerical instabilities, the results can differ between Theano and Tensorflow. However, as shown in \reftable{table:theano_vs_tensorflow}, we only see an insignificant difference between these two options. For Theano and Tensorflow we experience differences in terms of the runtime: Theano converts the computation graph first to C code and then runs a C compiler while Tensorflow links the operations of the computation graph against a pre-compiled build. Especially for complex networks Theano spends  a huge amount of time to compile the compute graph, which is sometimes the factor 10 - 30 times higher than the time required to run a single epoch. However, Theano implements several caching mechanism that help to some extent if the same architecture is re-executed at a later stage.   

Tensorflow started the training much faster, as the computation graph had only to be linked against pre-build functions. However, each trainings epoch took longer to run on a CPU, as the computation graph was not that well optimized. Therefore, in our specific case, there was no clear winner between the two backends in terms of run time.

\begin{table}[h]
\centering
\begin{tabular}{|c|c|c|c|c|}
\hline
\textbf{Task} & \textbf{\# Configs} & \textbf{Theano} & \textbf{Tensorflow}   \\ \hline
POS & 44 & \textbf{56.8\%} & 43.2\% \\
$\Delta Acc.$ &  &  & (-0.02\%) \\ \hline 
Chunking & 51 & \textbf{54.9\%} & 45.1\% \\
$\Delta F_1$ &  &  & (-0.02\%) \\ \hline 
NER & 65 & \textbf{50.8\%} & 49.2\% \\
$\Delta F_1$ &  &  & (-0.01\%) \\ \hline 
Entities & 55 & 49.1\% & \textbf{50.9\%} \\
$\Delta F_1$ &  & (-0.01\%) &  \\ \hline 
Events & 76 & 38.2\% & \textbf{61.8\%} \\
$\Delta F_1$ &  & (-0.12\%) &  \\ \hline 
\hline
Average &  & 49.9\% & \textbf{50.1\%} \\ \hline 
\end{tabular}
\caption{Network configurations were sampled randomly and each was evaluated with each possible backend. The first number in a cell depicts in how many cases each backend produced better results than the other. The second number shows the median difference to the best option for each task.}
\label{table:theano_vs_tensorflow}
\end{table}

\newpage
\section{Multi-Task Learning}\label{sec:mtl}
Multi-Task Learning has a long tradition in machine learning for neural networks \cite{Caruana1997}. In a multi-task learning (MTL) setup, the model is trained jointly for all tasks. The goal is to derive weights and representations that generalize well for a larger number of tasks and to achieve a better generalization on unseen data. 

The results for multi-task learning for NLP are so far mixed. \newcite{Collobert_2011} experimented with a feed-forward network for the task of POS tagging, Chunking and NER. They could achieve an improvement only for the Chunking task. \newcite{Sogaard2016} evaluated deep BiLSTM-networks and achieved an improvement for Chunking and CCG supertagging, while for NER, super senses (SemCor), and MWE brackets \& super sense tags no improvement was achieved. \newcite{Alonso2017} achieved only for 1 out of 5 tasks significant improvements.

In the following section, we evaluate the five sequence tagging tasks (POS, Chunking, NER, Entity Recognition and Event Recognition) in a multi-task learning scenario and analyze in which scenario an improvement can be achieved.

\subsection{Setup}
For our multi-task learning experiments share the different tasks the embedding layer as well as all BiLSTM-layers. To enable the network to output labels for more than one task, we add task specific output layers for each task. These are either softmax classifiers or CRF classifiers as described in \refsec{sec:lstm_model}. Training is achieved by minimizing the loss averaged across all tasks. As the training examples do not necessarily overlap, it is not possible to input a sentence and to compute the average loss across all tasks. Instead, we achieve this by picking alternatively examples for each task and apply the optimization step to all parameters of that task including the shared parameters  \cite{Collobert_2011}. 

For our specific experiment, we define one task as the main task and we will add another task as auxiliary task, for example the network is trained on the Chunking task with POS as auxiliary task. One training epoch is defined as one iteration over all training examples for the main task. When the auxiliary task has less training data then the main task, the same auxiliary training data will repetitively be used. When it has more, only a random fraction of it will be used per epoch. This gives each task equal weight. We will use the epoch with the highest development score for the main task. 

We then sampled 196 random network configurations. In contrast to the Single Task experiments, we restricted the tagging scheme to \textit{BIO}, the optimizer to \textit{Adam}, and the dropout type to \textit{variational dropout}. We evaluated each network configuration in a single task learning setup as well as adding one of the other datasets as auxiliary task. As described in  \refsec{sec:benchmark_tasks} only the first 500 sentences for the POS task are used if it is the main task. When it is used as auxiliary task, we use the full training set.

\subsection{Multi-Task Learning supervised at the same level}\label{sec:mtl_same_level}
\reftable{table:MTLvsSTL} depicts the result for the Single Task Learning (STL) vs.\ Multi-Task Learning experiment. For the chunking task, we see a clear outperformance when adding POS as auxiliary task ($p < 10^{-11}$) This is in line with previous observations from \newcite{Collobert_2011} and \newcite{Sogaard2016}. For the NER task, the Single Task Setup results in the best performance. Interestingly, the fairly similar Entities dataset did not help to improve the performance. For the Entities dataset, the Single Task Learning approach as well as adding the NER dataset performs on par. For the Events dataset, we see a clear performance increase when adding either the POS-dataset or the Chunking dataset.

\begin{table}[h]
\centering
\begin{tabular}{|c|c||c|c|c|c|c|c|}
\cline{3-7}
\multicolumn{2}{c|}{} & \multicolumn{5}{c|}{\bfseries Auxiliary Task } \\ \hline
\textbf{Main Task} & \textbf{STL} & \textbf{POS} & \textbf{Chunking} & \textbf{NER} & \textbf{Entities} & \textbf{Events}    \\ \hline
POS & 36.7\% &  & \textbf{57.7\%}$\dagger$ & 1.0\% & 1.0\% & 3.6\% \\
$\Delta Acc.$ & -0.08\% &  &  & -0.32\% & -0.54\% & -0.26\% \\  
$\sigma$ & 0.0060$\dagger$ &  & 0.0121 & 0.0090 & 0.0178 & 0.0105 \\ \hline 
Chunking &  25.5\% & \textbf{74.0\%}$\dagger$ & & 0.5\% & 0.0\% & 0.0\% \\
$\Delta F_1$ & -0.22\%  &  &  & -0.45\% & -0.53\% & -0.62\% \\  
$\sigma$ & 0.0040 & 0.0041 &  & 0.0046 & 0.0047 & 0.0045 \\ \hline 
NER & \textbf{65.8\%}$\dagger$ & 12.8\% & 0.0\% &  & 17.9\% & 3.6\% \\
$\Delta F_1$ &  & -0.50\% & -0.89\% &  & -0.38\% & -0.76\% \\  
$\sigma$ & 0.0057$\dagger$ & 0.0067$\dagger$ & 0.0079 &  & 0.0074 & 0.0071 \\ \hline 
Entities & 46.4\%$\dagger$ & 1.0\% & 0.5\% & \textbf{51.5\%}$\dagger$ &  & 0.5\% \\
$\Delta F_1$ & -0.09\% & -1.25\% & -1.86\% &  &  & -1.71\% \\  
$\sigma$ & 0.0091$\dagger$ & 0.0104$\dagger$ & 0.0121 & 0.0103$\dagger$ &  & 0.0143 \\ \hline 
Events &  4.1\% & \textbf{44.4\%}$\dagger$ & 40.8\%$\dagger$ & 4.6\% & 6.1\% & \\
$\Delta F_1$ & -0.64\% &  & -0.01\% & -0.49\% & -0.58\% &  \\  
$\sigma$ & 0.0047$\dagger$  & 0.0046$\dagger$ & 0.0043$\dagger$ & 0.0052 & 0.0055 & \\ \hline 
\hline
Average &  & \textbf{33.8\%} & 24.9\% & 24.7\% & 14.3\% & 2.3\% \\ \hline 
\end{tabular}
\caption{Single Task Learning (STL) vs. Multi-Task Learning with different auxiliary datasets. For each task, 196 randomly sampled configurations were evaluated. First, in a STL setup and then with each other dataset as auxiliary dataset. The first number depicts in how many cases each setup resulted in the best test performance. The second number shows the median difference to the best option for each task. The third number the standard deviation $\sigma$ of observed test scores. Statistically significant differences with $p < 0.01$ are marked with $\dagger$.  }
\label{table:MTLvsSTL}
\end{table}

\begin{figure}[ht]
\centering
  \includegraphics[width=0.8\textwidth]{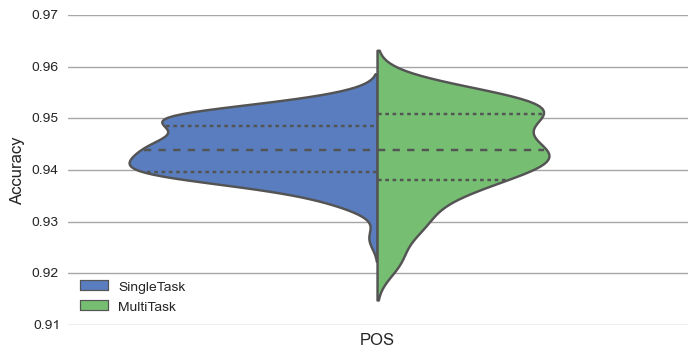}
\caption{Comparison of the performance on the POS dataset. Blue/Left: Single Task Learning setup. Green/Right: Multi-Task Learning setup with Chunking as auxiliary data.}
\label{fig:mtl_pos_chunking}
\end{figure}

For the POS-dataset, we get a statistically significant outperformance when using Chunking as an additional task. However, the median performance difference is with -0.08\% fairly small. \reffigure{fig:mtl_pos_chunking} depicts the probability density of these two setups. As the figure shows, we observe a larger variance for the multi-task setup: We see several ill performing configurations, but also several configurations that clearly outperform the single task setup. The standard deviation $\sigma$ in \reftable{table:MTLvsSTL} confirms that that the single task setup has the lowest variance for all tasks and setups. 

We conclude that hyperparameter testing is more important for the multi-task setup than it is in a single-task setup. Finding a well working hyperparameter configuration and finding a stable local minimum that generalizes well to unseen data is more challenging for the multi-task setup than it is for the single-task setup.

\FloatBarrier
\subsection{Supervising tasks at different levels}

\newcite{Sogaard2016} present the idea to supervise low level tasks at lower layers. For example a stacked BiLSTM network with three layers, the output of the first LSTM layer could be used to classify part-of-speech tags and the output of the third layer to classify chunking tags. In their paper, they could observe a performance improvement for chunking as well as for CCG supertagging.

In order to evaluate this idea, we sampled 188 distinct network configurations. For each configuration we required three stacked BiLSTM layers, each layer with a randomly selected number of recurrent units in the set $\{25,50,75,100,125\}$. We only evaluated combinations that showed in \refsec{sec:mtl_same_level} that multi-task learning helps to improve performance. For each main task, we evaluated it in a Single Task setup as well as with one auxiliary task supervised at either the same output level or at different levels (main task at the thrid, auxiliary task at the first level). \reftable{table:MTLvsSTL_output_level} shows the results of this experiment. 

\begin{table}[h]
\centering
\begin{tabular}{|c|c|c|c|c|c|}
\hline
\textbf{Main Task} & \textbf{Aux. Task} & \textbf{\# Configs} & \textbf{Single Task} & \textbf{Same Level} & \textbf{Different Level}    \\ \hline
POS & Chunking & 188 & 10.6\% & 32.4\% & \textbf{56.9\%}$\dagger$ \\
$\Delta Acc.$ & &  & -0.67\% & -0.19\% &  \\  
$\sigma$ &  & & 0.0191 & 0.0217 & 0.0191 \\ \hline 
Chunking & POS & 188 & 16.0\% & 24.5\% & \textbf{59.6\%}$\dagger$ \\
$\Delta F_1$ &  & & -0.36\% & -0.12\% &  \\  
$\sigma$ &  & & 0.0065 & 0.0079 & 0.0064 \\ \hline 
Entities & NER & 188 & 21.8\% & 35.1\%$\dagger$ & \textbf{43.1\%}$\dagger$ \\
$\Delta F_1$ &  & & -0.36\% & -0.17\% &  \\  
$\sigma$ &  & & 0.0138 & 0.0161 & 0.0132 \\ \hline 
Events & POS & 188 & 5.9\% & \textbf{56.4\%}$\dagger$ & 37.8\%$\dagger$ \\
$\Delta F_1$ & & & -0.90\% &  & -0.12\% \\  
$\sigma$ &  & & 0.0143 & 0.0111$\dagger$ & 0.0124$\dagger$ \\ \hline 
\hline
Average & &  & 13.6\% & 37.1\% & \textbf{49.3\%} \\ \hline 
\end{tabular}
\caption{188 configurations for the BiLSTM sequence tagging model were sampled at random and evaluated in a Single Task Learning setup, in a Multi-Task Learning setup with two tasks supervised at the same output level, and in a Multi-Task Learning setup with the main task supervised at the third level and the auxiliary task supervised at the first level.}
\label{table:MTLvsSTL_output_level}
\end{table}

As the results show, supervising at different levels usually improves the performance. Only for the Events dataset supervising at the same output level would be the best option. However, the performance difference with 0.12 percentage points  is rather small. 

In contrast to \newcite{Sogaard2016} we observe that not only low level tasks should be supervised at lower layers. We could observe a performance increase when the POS task was supervised at the third layer and Chunking at the first layer as well as vice versa, when POS was supervised at the first and Chunking at the third layer. It appears that it is beneficial to have LSTM layers that are optimized specifically for a single task, i.e.\ the split between the models for the different tasks should happen before the last layer and each task should have task-dependent LSTM-layers. 

A frequently missing point in the evaluation of multi-task learning is the comparison to a pipeline approach: Often, only one task, usually defined as the main task, can be improved by multi-task learning. In such a case, a pipeline approach, where the output labels of the auxiliary tasks are added as features to the main task might, outperform the multi-task setup. We leave this evaluation of multi-task learning versus pipeline approach as feature work.

\section{Conclusion}
In this work, we evaluated different design choices and hyperparameters for the commonly used BiLSTM-architecture for linguistic sequence tagging. As \reftable{table:random_init} showed, the random initialization of the weights has a major impact on the test performance. Leaving all hyperparameters the same, just changing the seed value of the random number generator, led to average test performance differences of 0.72 percentage points for the ACE 2005 entities recognition task. This requires that a certain configuration of the network is evaluated multiple times in order to draw conclusions. In order to cancel out random noise, we evaluated \numRuns{} network configurations to draw conclusions which parameters yield the best performance. 

Our results in \refsec{sec:evaluation_results} reveal that the choice of word embeddings, the selected optimizer, the classifier, used as last layer, and the dropout mechanism have a high impact on the achieved performance. We showed that the embeddings by \newcite{Komninos2016} typically perform best, that Adam with Nesterov momentum (Nadam) \cite{Nadam} results in the best performance and converges the fastest, that a gradient normalization threshold of 1 should be used, that variational dropout \cite{Gal2015} applied to the output units as well as the recurrent units of an LSTM-layer is optimal and that the CRF-classifier presented \cite{Huang2015} should be preferred over a softmax classifier. Other design choices, for example the type of character representation, the tagging scheme, the number of LSTM layers and the number of recurrent units only had a minor impact for the evaluated tasks.

In \refsec{sec:mtl} we evaluated different multi-task learning setups. While the majority of combinations of tasks does not result in a performance increase, we can see for some combinations a consistent performance increase, especially for tasks that are fairly similar. Further we could observe that the performance variance for multi-task learning is higher than for single task learning. We conclude that multi-task learning is especially sensitive to the selection of hyperparameters and finding local minima with low generalization error appears more challenging than for single task learning.

Further, we evaluated the supervision of tasks at different levels as described by \newcite{Sogaard2016} and observed, that not only low level tasks profit from supervision at lower layers, but that it appears to be in general superficial to have task-specific LSTM-layers besides shared LSTM-layers. 

During our experiments, we usually only looked at one dimension for a certain hyperparameter, e.g.\ which pre-trained word embedding results in the best performance. However, hyperparameters can influence each other and the individual best options must not necessarily be the global optimum. We observed this when studying the optimal number of recurrent units, which yield different solutions depending on the number of recurrent layers.

\section*{Acknowledgments}
This work has been supported by the German Research Foundation as part of the Research Training Group ``Adaptive Preparation of Information from Heterogeneous Sources'' (AIPHES) under grant No.\ GRK 1994/1. Calculations for this research were conducted on the Lichtenberg high performance computer of the TU Darmstadt.

\FloatBarrier
\bibliographystyle{acl}
\bibliography{references}

\end{document}